\documentstyle[jair,twoside,11pt,theapa,epsfig,amssymb]{article}
\input epsf 

\newtheorem{theorem}{Theorem}[section]
\newtheorem{lemma}[theorem]{Lemma}
\newtheorem{definition}[theorem]{Definition}
\newtheorem{proposition}[theorem]{Proposition}

\def\proof{\noindent{\bf Proof }}
\def\endproof{\hspace*{\fill}\par\endtrivlist\unskip}

\def\THREES{\textsc{3S}}
\def\PECn{\textsc{Plan-Existence-}${\mathbb C}_n${}}
\def\CNFSAT{\textsc{Cnf-Sat}}
\def\THREESAT{\textsc{3Sat}}

\jairheading{31}{2008}{319-351}{09/07}{02/08}
\ShortHeadings{Complexity of Planning Problems}{Gim\'enez \& Jonsson}
\firstpageno{319}

\begin{document}

\title{The Complexity of Planning Problems\\ With Simple Causal Graphs}

\author{\name Omer Gim\'enez \email omer.gimenez@upc.edu\\
	\addr Dept. de Llenguatges i Sistemes Inform\`atics\\
	Universitat Polit\`ecnica de Catalunya\\
	Jordi Girona, 1-3\\
	08034 Barcelona, Spain
	\AND
	\name Anders Jonsson \email anders.jonsson@upf.edu\\
	\addr Dept. of Information and Communication Technologies\\
	Passeig de Circumval$\cdot$laci\'o, 8\\
	08003 Barcelona, Spain}

\maketitle

\begin{abstract}
We present three new complexity results for classes of planning
problems with simple causal graphs.  First, we describe a
polynomial-time algorithm that uses macros to generate plans for the
class \THREES{} of planning problems with binary state variables and
acyclic causal graphs.  This implies that plan generation may be
tractable even when a planning problem has an exponentially long
minimal solution.  We also prove that the problem of plan existence
for planning problems with multi-valued variables and chain causal
graphs is $\mathrm{NP}$-hard.  Finally, we show that plan existence
for planning problems with binary state variables and polytree causal
graphs is $\mathrm{NP}$-complete.
\end{abstract}

\section{Introduction}

Planning is an area of research in artificial intelligence that aims
to achieve autonomous control of complex systems.  Formally, the
planning problem is to obtain a sequence of transformations for moving
a system from an initial state to a goal state, given a description of
possible transformations.  Planning algorithms have been successfully
used in a variety of applications, including robotics, process
planning, information gathering, autonomous agents and spacecraft
mission control. Research in planning has seen significant progress
during the last ten years, in part due to the establishment of the
International Planning Competition.

An important aspect of research in planning is to classify the
complexity of solving planning problems. Being able to classify a
planning problem according to complexity makes it possible to select
the right tool for solving it. Researchers usually distinguish between
two problems: plan generation, the problem of generating a sequence of
transformations for achieving the goal, and plan existence, the
problem of determining whether such a sequence exists. If the original
STRIPS formalism is used, plan existence is undecidable in the
first-order case \cite{Chapman87} and PSPACE-complete in the
propositional case \cite{Bylander94}. Using PDDL, the representation
language used at the International Planning Competition, plan
existence is EXPSPACE-complete \cite{Erol95}.  However, planning
problems usually exhibit structure that makes them much easier to
solve. \citeA{Helmert03} showed that many of the benchmark problems
used at the International Planning Competition are in fact in
$\mathrm{P}$ or $\mathrm{NP}$.

A common type of structure that researchers have used to characterize
planning problems is the so called \emph{causal graph}
\cite{Knoblock94}. The causal graph of a planning problem is a graph
that captures the degree of independence among the state variables of
the problem, and is easily constructed given a description of the
problem transformations. The independence between state variables can
be exploited to devise algorithms for efficiently solving the planning
problem. The causal graph has been used as a tool for describing
tractable subclasses of planning problems
\cite{Brafman03,Jonsson98b,Williams97}, for decomposing planning
problems into smaller problems \cite{Brafman06,Jonsson07,Knoblock94},
and as the basis for domain-independent heuristics that guide the
search for a valid plan \cite{Helmert06b}.

In the present work we explore the computational complexity of
solving planning problems with simple causal graphs. We present
new results for three classes of planning problems studied in
the literature: the class \THREES{} \cite{Jonsson98b}, the class
${\mathbb C}_n$ \cite{Domshlak01}, and the class of planning
problems with polytree causal graphs \cite{Brafman03}.
In brief, we show that plan generation for instances of the
first class can be solved in polynomial time using macros, but
that plan existence is not solvable in polynomial time for the
remaining two classes, unless $\mathrm{P}=\mathrm{NP}$. This
work first appeared in a conference paper \cite{Gimenez07}; the
current paper provides more detail and additional insights as
well as new sections on plan length and CP-nets.

A planning problem belongs to the class \THREES{} if its causal graph
is acyclic and all state variables are either \emph{static},
\emph{symmetrically reversible} or \emph{splitting} (see
Section~\ref{section:3S} for a precise definition of these terms).
The class \THREES{} was introduced and studied by \citeA{Jonsson98b}
as an example of a class for which plan existence is easy (there
exists a polynomial-time algorithm that determines whether or not a
particular planning problem of that class is solvable) but plan
generation is hard (there exists no polynomial-time algorithm that
generates a valid plan for every planning problem of the class). More
precisely, \citeauthor{Jonsson98b} showed that there are planning problems
of the class \THREES{} for which every valid plan is exponentially
long. This clearly prevents the existence of an efficient plan
generation algorithm.

Our first contribution is to show that plan generation for \THREES{}
is in fact easy if we are allowed to express a valid plan using
\emph{macros}. A macro is simply a sequence of operators and
other macros. We present a polynomial-time algorithm that produces
valid plans of this form for planning problems of the class \THREES.
Namely, our algorithm outputs in polynomial time a system of macros
that, when executed, produce the actual valid plan for the planning
problem instance. The algorithm is sound and complete, that is, it
generates a valid plan if and only if one exists. We contrast our
algorithm to the incremental algorithm proposed by \citeA{Jonsson98b},
which is polynomial in the size of the output.

We also investigate the complexity of the class ${\mathbb C}_n$ of
planning problems with multi-valued state variables and chain causal
graphs.  In other words, the causal graph is just a directed path.
\citeA{Domshlak01} showed that there are solvable instances of this
class that require exponentially long plans. However, as it is the
case with the class \THREES, there could exist an efficient procedure
for generating valid plans for ${\mathbb C}_n$ instances using macros
or some other novel idea. We show that plan existence in ${\mathbb
C}_n$ is $\mathrm{NP}$-hard, hence ruling out that such an efficient
procedure exists, unless $\mathrm{P}=\mathrm{NP}$.

We also prove that plan existence for planning problems whose causal
graph is a polytree (i.e., the underlying undirected graph is acyclic)
is $\mathrm{NP}$-complete, even if we restrict to problems with binary
variables. This result closes the complexity gap that appears in
\citeA{Brafman03} regarding planning problems with binary variables.
The authors show that plan existence is $\mathrm{NP}$-complete for
planning problems with singly connected causal graphs, and that plan
generation is polynomial for planning problems with polytree causal
graphs of bounded indegree.  We use the same reduction to prove that a
similar problem on polytree CP-nets \cite{Boutilier04} is
$\mathrm{NP}$-complete.

\subsection{Related Work}

Several researchers have used the causal graph to devise algorithms
for solving planning problems or to study the complexity of planning
problems. \citeA{Knoblock94} used the causal graph to decompose a
planning problem into a hierarchy of increasingly abstract problems.
Under certain conditions, solving the hierarchy of abstract problems
is easier than solving the original problem. \citeA{Williams97}
introduced several restrictions on planning problems to ensure
tractability, one of which is that the causal graph should be
acyclic. \citeA{Jonsson98b} defined the class \THREES{} of planning
problems, which also requires the causal graphs to be acyclic, and
showed that plan existence is polynomial for this class.

\citeA{Domshlak01} analyzed the complexity of several classes of
planning problems with acyclic causal graphs. \citeA{Brafman03} 
designed a polynomial-time algorithm for solving planning problems
with binary state variables and acyclic causal graph of bounded
indegree. \citeA{Brafman06} identified conditions under which it is
possible to factorize a planning problem into several subproblems
and solve the subproblems independently. They claimed that a planning
problem is suitable for factorization if its causal graph has
bounded tree-width.

The idea of using macros in planning is almost as old as planning
itself \cite{Fikes71}. \citeA{Minton85} developed an algorithm that
measures the utility of plan fragments and stores them as macros if
they are deemed useful. \citeA{Korf87} showed that macros can
exponentially reduce the search space size of a planning problem if
chosen carefully. \citeA{Vidal04} used relaxed plans generated while
computing heuristics to produce macros that contribute to the
solution of planning problems. Macro-FF \cite{Botea05}, an algorithm
that identifies and caches macros, competed at the fourth
International Planning Competition. The authors showed how macros
can help reduce the search effort necessary to generate a valid
plan.

\citeA{Jonsson07} described an algorithm that uses macros
to generate plans for planning problems with tree-reducible
causal graphs. There exist planning problems for which the
algorithm can generate exponentially long solutions in polynomial
time, just like our algorithm for \THREES. Unlike ours, the algorithm
can handle multi-valued variables, which enables it to solve
problems such as Towers of Hanoi. However, not all planning problems
in \THREES{} have tree-reducible causal graphs, so the algorithm cannot be
used to show that plan generation for \THREES{} is polynomial.

\subsection{Hardness and Plan Length}

A contribution of this paper is to show that plan generation may be
polynomial even when planning problems have exponential length minimal
solutions, provided that solutions may be expressed using a concise
notation such as macros.  We motivate this result below and discuss
the consequences.  Previously, it has been thought that plan
generation for planning problems with exponential length minimal
solutions is harder than $\mathrm{NP}$, since it is not known whether
problems in $\mathrm{NP}$ are intractable, but it is certain that we
cannot generate exponential length output in polynomial time.

However, for a planning problem with exponential length minimal
solution, it is not clear if plan generation is inherently hard, or if
the difficulty just lies in the fact that the plan is long.  Consider
the two functional problems
$$ f_1(F) = w({\tt 1}, 2^{|F|}),
$$ $$ f_2(F) = w(t(F), 2^{|F|}), $$
where $F$ is a $3$-CNF formula, $|F|$ is the number of clauses of $F$,
$w(\sigma, k)$ is a word containing $k$ copies of the symbol $\sigma$,
and $t(F)$ is $\tt 1$ if $F$ is satisfiable (i.e., $F$ is in \THREESAT), and
$\tt 0$ if it is not.  In both cases, the problem consists in generating
the correct word.  Observe that both $f_1$ and $f_2$ are provably
intractable, since their output is exponential in the size of the
input.

Nevertheless, it is intuitive to regard problem $f_1$ as \emph{easier}
than problem $f_2$. One way to formalize this intuition is to allow
programs to produce the output in some succinct notation. For
instance, if we allow programs to write ``${\tt w(\sigma, }k{\tt )}$'' instead
of a string containing $k$ copies of the symbol $\sigma$, then problem
$f_1$ becomes polynomial, but problem $f_2$ does not (unless
$\mathrm{P}=\mathrm{NP}$).

We wanted to investigate the following question: regarding the class
\THREES, is plan generation intractable because solution plans are long,
like $f_1$, or because the problem is intrinsically hard, like $f_2$?
The answer is that plan generation for \THREES{} can be solved in
polynomial time, provided that one is allowed to give the solution in
terms of macros, where a macro is a simple substitution scheme: a
sequence of operators and/or other macros.  To back up this claim, we
present an algorithm that solves plan generation for \THREES{} in
polynomial time.

Other researchers have argued intractability using the fact that plans
may have exponential length. \citeA{Domshlak01} proved complexity
results for several classes of planning problems with multi-valued
state variables and simple causal graphs.  They argued that the class
${\mathbb C}_n$ of planning problems with chain causal graphs is
intractable since plans may have exponential length.
\citeA{Brafman03} stated that plan generation for STRIPS planning
problems with unary operators and acyclic causal graphs is intractable
using the same reasoning.  Our new result puts in question the
argument used to prove the hardness of these problems. For this
reason, we analyze the complexity of these problems and prove that
they are hard by showing that the plan existence problem is
$\mathrm{NP}$-hard.

\section{Notation}

Let $V$ be a set of state variables, and let $D(v)$ be the finite
domain of state variable $v \in V$.  We define a state $s$ as a
function on $V$ that maps each state variable $v \in V$ to a value
$s(v) \in D(v)$ in its domain.  A partial state $p$ is a function on a
subset $V_p \subseteq V$ of state variables that maps each state
variable $v \in V_p$ to $p(v) \in D(v)$.  For a subset $C \subseteq V$
of state variables, $p \mid C$ is the partial state obtained by
restricting the domain of $p$ to $V_p \cap C$. Sometimes we use the
notation $(v_1 = x_1, \ldots, v_k = x_k)$ to denote a partial state
$p$ defined by $V_p = \{v_1, \ldots, v_k\}$ and $p(v_i) = x_i$ for
each $v_i \in V_p$. We write $p(v) = \perp$ to denote that $v \notin V_p$.

Two partial states $p$ and $q$ match, which we denote $p \triangledown
q$, if and only if $p \mid V_q = q \mid V_p$, i.e., for each $v \in
V_p \cap V_q$, $p(v) = q(v)$. We define a replacement operator
$\oplus$ such that if $q$ and $r$ are two partial states, $p = q
\oplus r$ is the partial state defined by $V_p = V_q \cup V_r$, $p(v)
= r(v)$ for each $v \in V_r$, and $p(v) = q(v)$ for each $v \in V_q -
V_r$. Note that, in general, $p \oplus q \neq q \oplus p$. A partial
state $p$ subsumes a partial state $q$, which we denote $p \sqsubseteq
q$, if and only if $p \triangledown q$ and $V_p \subseteq V_q$. We remark
that if $p \sqsubseteq q$ and $r \sqsubseteq s$, it follows that $p
\oplus r \sqsubseteq q \oplus s$.  The difference between two partial
states $q$ and $r$, which we denote $q - r$, is the partial state $p$
defined by $V_p = \{v \in V_q \mid q(v) \neq r(v)\}$ and $p(v) = q(v)$
for each $v \in V_p$.

A planning problem is a tuple $P = \langle V, init, goal, A \rangle$,
where $V$ is the set of variables, $init$ is an initial state, $goal$
is a partial goal state, and $A$ is a set of operators.  An operator
$a = \langle pre(a); post(a) \rangle \in A$ consists of a partial
state $pre(a)$ called the {\em pre-condition} and a partial state
$post(a)$ called the {\em post-condition}.  Operator $a$ is applicable
in any state $s$ such that $s \triangledown pre(a)$, and applying
operator $a$ in state $s$ results in the new state $s \oplus post(a)$.
A valid plan $\Pi$ for $P$ is a sequence of operators that are
sequentially applicable in state $init$ such that the resulting state
$s'$ satisfies $s' \triangledown goal$.

The causal graph of a planning problem $P$ is a directed graph $(V,
E)$ with state variables as nodes.  There is an edge $(u, v) \in E$ if
and only if $u \neq v$ and there exists an operator $a \in A$ such
that $u \in V_{pre(a)} \cup V_{post(a)}$ and $v \in V_{post(a)}$.

\section{The Class \THREES} \label{section:3S}

\citeA{Jonsson98b} introduced the class \THREES{} of planning problems
to study the relative complexity of plan existence and plan
generation.  In this section, we introduce additional notation needed
to describe the class \THREES{} and illustrate some of the properties
of \THREES{} planning problems. We begin by defining the class
\THREES:

\begin{definition}
A planning problem $P$ belongs to the class \THREES{} if its causal
graph is acyclic and each state variable $v \in V$ is binary and
either static, symmetrically reversible, or splitting.
\end{definition}

Below, we provide formal definitions of {\em static}, {\em
symmetrically reversible} and {\em splitting}. Note that the fact that
the causal graph is acyclic implies that operators are unary, i.e.,
for each operator $a\in A$, $|V_{post(a)}| = 1$. Without loss of
generality, we assume that \THREES{} planning problems are in {\em
normal form}, by which we mean the following:

\begin{itemize}
\item For each state variable $v$, $D(v)=\{0,1\}$ and $init(v)=0$.
\item $post(a)=(v=x)$, $x \in \{0,1\}$, implies that $pre(a)(v)=1-x$.
\end{itemize}

To satisfy the first condition, we can relabel the values of $D(v)$ in
the initial and goal states as well as in the pre- and post-conditions
of operators. To satisfy the second condition, for any operator $a$
with $post(a)=(v=x)$ and $pre(a)(v)\neq 1-x$, we either remove it if
$pre(a)(v)=x$, or we let $pre(a)(v)=1-x$ if previously undefined.  The
resulting planning problem is in normal form and is equivalent to the
original one. This process can be done in time $O(|A||V|)$.

The following definitions describe the three categories of state
variables in \THREES:

\begin{definition}
A state variable $v \in V$ is {\em static} if and only if one of the
following holds:
\end{definition}
\begin{enumerate}
\item {\it There does not exist $a \in A$ such that $post(a)(v) = 1$,}
\item {\it $goal(v) = 0$ and there does not exist $a \in A$ such that
$post(a)(v) = 0$.}
\end{enumerate}

\begin{definition}
A state variable $v \in V$ is {\em reversible} if and only if for each
$a \in A$ such that $post(a)=(v=x)$, there exists $a' \in A$ such that
$post(a')=(v=1-x)$.  In addition, $v$ is {\em symmetrically reversible}
if $pre(a') \mid (V - \{v\}) = pre(a) \mid (V - \{v\})$.
\end{definition}

From the above definitions it follows that the value of a static state
variable cannot or must not change, whereas the value of a
symmetrically reversible state variable can change freely, as long as
it is possible to satisfy the pre-conditions of operators that change
its value.  The third category of state variables is {\em
splitting}. Informally, a splitting state variable $v$ splits the
causal graph into three disjoint subgraphs, one which depends on the
value $v = 1$, one which depends on $v = 0$, and one which is
independent of $v$. However, the precise definition is more involved,
so we need some additional notation.

For $v \in V$, let $Q_0^v$ be the subset of state variables, different
from $v$, whose value is changed by some operator that has $v = 0$ as
a pre-condition.  Formally, $Q_0^v = \{u \in V - \{v\} \mid \exists a
\in A \; {\mathrm s.t.} \; pre(a)(v) = 0 \wedge u \in V_{post(a)}\}$.
Define $Q_1^v$ in the same way for $v = 1$. Let $G_0^v = (V, E_0^v)$
be the subgraph of $(V, E)$ whose edges exclude those between $v$ and
$Q_0^v - Q_1^v$. Formally, $E_0^v = E - \{(v, w) \mid w \in Q_0^v
\wedge w \notin Q_1^v\}$.  Finally, let $V_0^v \subseteq V$ be the
subset of state variables that are weakly connected to some state
variable of $Q_0^v$ in the graph $G_0^v$. Define $V_1^v$ in the same
way for $v = 1$.

\begin{definition}
A state variable $v \in V$ is splitting if and only if $V_0^v$ and
$V_1^v$ are disjoint.
\end{definition}

\begin{figure}
\centerline{\epsfig{figure=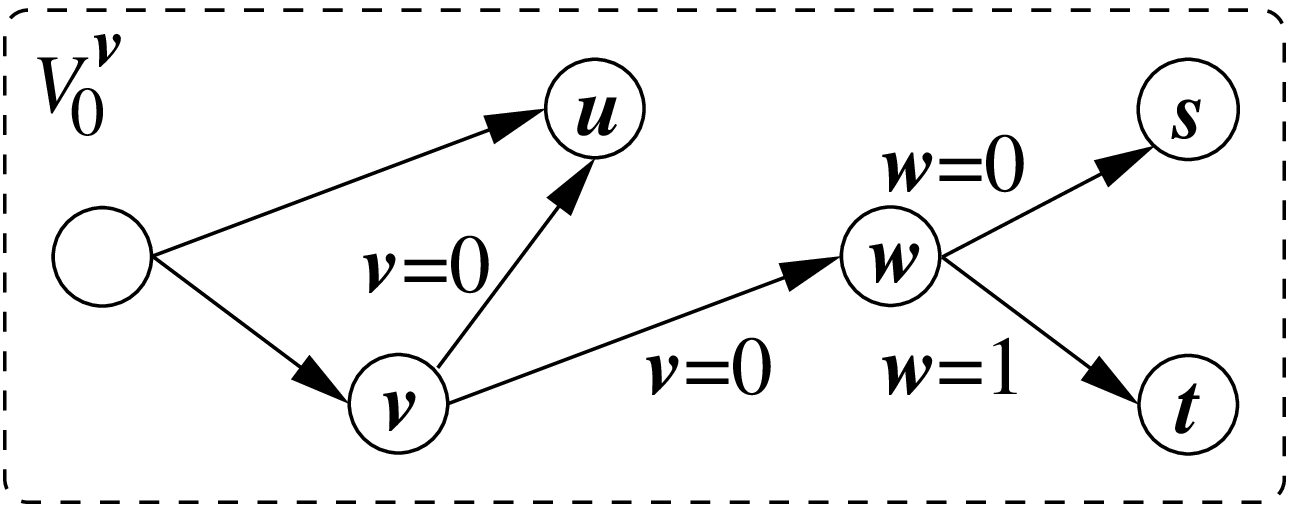,height=2.5cm}
\hspace{1cm}
\epsfig{figure=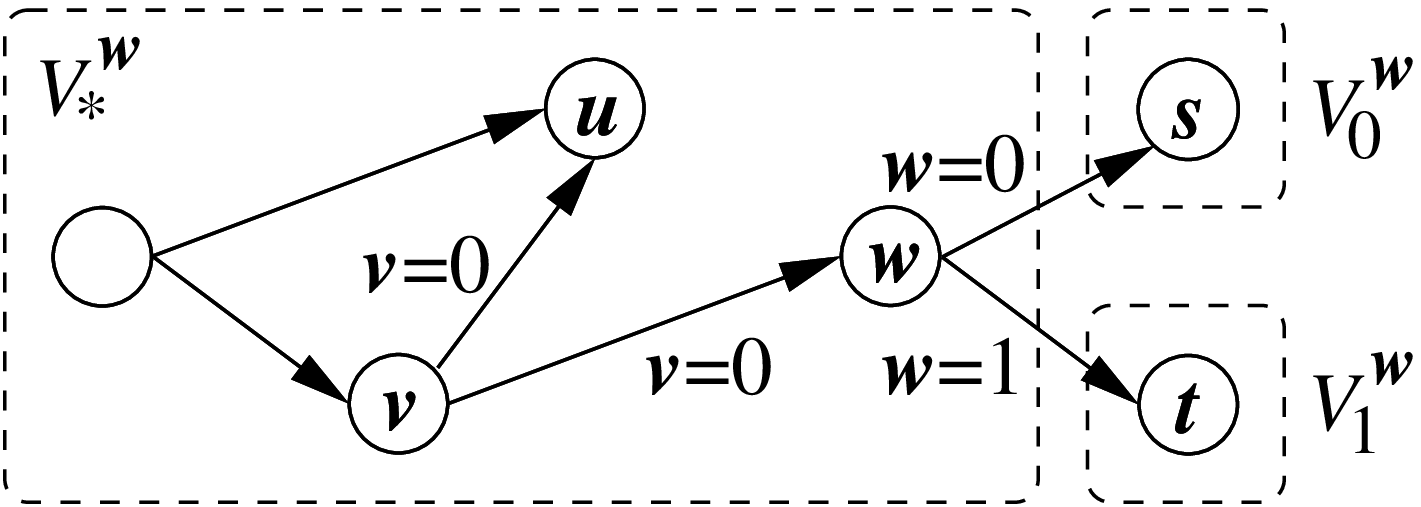,height=2.5cm}}
\centerline{(a) \hspace{7cm} (b)}
\vspace{16pt}
\caption{Causal graph with splitting variable partitions for (a) $v$, (b) $w$.}
\label{fig:causalsplitting}
\end{figure}

Figure~\ref{fig:causalsplitting} illustrates the causal graph of a
planning problem with two splitting state variables, $v$ and
$w$. The edge label $v = 0$ indicates that there are operators for
changing the value of $u$ that have $v = 0$ as a pre-condition. In
other words, $Q_0^v = \{u, w\}$, the graph $G_0^v = (V, E_0^v)$
excludes the two edges labeled $v = 0$, and $V_0^v$ includes all state
state variables, since $v$ is weakly connected to $u$ and $w$ connects
to the remaining state variables. The set $Q_1^v$ is empty since there
are no operators for changing the value of a state variable other than
$v$ with $v = 1$ as a pre-condition. Consequently, $V_1^v$ is empty as
well. Figure~\ref{fig:causalsplitting}(a) shows the resulting
partition for $v$.

In the case of $w$, $Q_0^w = \{s\}$, $G_0^w = (V, E_0^w)$ excludes the
edge labeled $w = 0$, and $V_0^w = \{s\}$, since no other state
variable is connected to $s$ when the edge $w = 0$ is removed.
Likewise, $V_1^w = \{t\}$. We use $V_*^w = V - V_0^w - V_1^w$ to
denote the set of remaining state variables that belong neither to
$V_0^w$ nor to $V_1^w$. Figure~\ref{fig:causalsplitting}(b) shows the
resulting partition for $w$.

\begin{lemma}
\label{lemma:nonempty}
For any splitting state variable $v$, if the two sets $V_0^v$ and
$V_1^v$ are non-empty, $v$ belongs neither to $V_0^v$ nor to $V_1^v$.
\end{lemma}

\proof
By contradiction. Assume that $v$ belongs to $V_0^v$. Then $v$ is
weakly connected to some state variable of $Q_0^v$ in the graph $G_0^v
= (V, E_0^v)$. But since $E_0^v$ does not exclude edges between $v$
and $Q_1^v$, any state variable in $Q_1^v$ is weakly connected to the
same state variable of $Q_0^v$ in $G_0^v$.  Consequently, state
variables in $Q_1^v$ belong to both $V_0^v$ and $V_1^v$, which
contradicts that $v$ is splitting. The same reasoning holds to show that $v$
does not belong to $V_1^v$.
\endproof

\begin{lemma}
The value of a splitting state variable never needs to change more
than twice on a valid plan.
\end{lemma}

\proof
Assume $\Pi$ is a valid plan that changes the value of a splitting
state variable $v$ at least three times. We show that we can reorder
the operators of $\Pi$ in such a way that the value of $v$ does not
need to change more than twice. We need to address three cases: $v$
belongs to $V_0^v$ (cf. Figure~\ref{fig:causalsplitting}(a)), $v$
belongs to $V_1^v$, or $v$ belongs to $V_*^v$
(cf. Figure~\ref{fig:causalsplitting}(b)).

If $v$ belongs to $V_0^v$, it follows from Lemma \ref{lemma:nonempty}
that $V_1^v$ is empty. Consequently, no operator in the plan requires
$v = 1$ as a pre-condition. Thus, we can safely remove all operators
in $\Pi$ that change the value of $v$, except possibly the last, which
is needed in case $goal(v) = 1$. If $v$ belongs to $V_1^v$, it follows
from Lemma \ref{lemma:nonempty} that $V_0^v$ is empty. Thus, no
operator in the plan requires $v = 0$ as a pre-condition. The first
operator in $\Pi$ that changes the value of $v$ is necessary to set
$v$ to 1. After that, we can safely remove all operators in $\Pi$ that
change the value of $v$, except the last in case $goal(v) = 0$. In
both cases the resulting plan contains at most two operators
changing the value of $v$.

If $v$ belongs to $V_*^v$, then the only edges between $V_0^v$,
$V_1^v$, and $V_*^v$ are those from $v\in V_*^v$ to $Q_0^v\subseteq
V_0^v$ and $Q_1^v\subseteq V_1^v$. Let $\Pi_0, \Pi_1$, and $\Pi_*$ be
the subsequences of operators in $\Pi$ that affect state variables in
$V_0^v, V_1^v$, and $V_*^v$, respectively. Write $\Pi_*=\langle
\Pi_*', a_1^v, \Pi_*''\rangle$, where $a^v_1$ is the last operator in
$\Pi_*$ that changes the value of $v$ from $0$ to $1$.
%
%
%
We claim that the reordering $\langle \Pi_0, \Pi_*', a_1^v, \Pi_1,
\Pi_*'' \rangle$ of plan $\Pi$ is still valid. Indeed, the operators
of $\Pi_0$ only require $v = 0$, which holds in the initial state, and
the operators of $\Pi_1$ only require $v = 1$, which holds due to the
operator $a_1^v$. Note that all operators changing the value of $v$ in
$\Pi_*'$ can be safely removed since the value $v = 1$ is never
needed as a pre-condition to change the value of a state variable in
$V_*^v$. The result is a valid plan that changes the value of $v$ at
most twice (its value may be reset to 0 by $\Pi_*''$).\\ \endproof

The previous lemma, which holds for splitting state variables in
general, provides some additional insight into how to solve
a planning problem with a splitting state variable $v$. First, try to
achieve the goal state for state variables in $V_0^v$ while the value
of $v$ is 0, as in the initial state. Then, set the value of $v$ to
$1$ and try to achieve the goal state for state variables in
$V_1^v$. Finally, if $goal(v)=0$, reset the value of $v$ to $0$.

\subsection{Example}

\begin{figure}
\centerline{\epsfig{figure=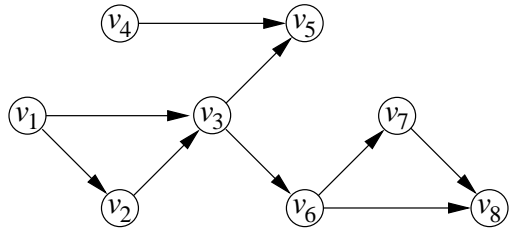,height=3cm}}
\vspace{16pt}
\caption{Causal graph of the example planning problem.}
\label{fig:3s}
\end{figure}

We illustrate the class \THREES{} using an example planning problem.
The set of state variables is $V = \{v_1, \ldots, v_8\}$. Since the
planning problem is in normal form, the initial state is $init(v_i) =
0$ for each $v_i \in V$. The goal state is defined by $goal = (v_5 =
1, v_8 = 1)$, and the operators in $A$ are listed in Table
\ref{table:ex}. Figure \ref{fig:3s} shows the causal graph $(V, E)$ of
the planning problem. From the operators it is easy to verify that
$v_4$ is static and that $v_1$ and $v_6$ are symmetrically
reversible. For the planning problem to be in \THREES, the remaining
state variables have to be splitting. Table \ref{table:ex} lists the
two sets $V_0^{v_i}$ and $V_1^{v_i}$ for each state variable $v_i \in
V$ to show that indeed, $V_0^{v_i} \cap V_1^{v_i} = \emptyset$ for
each of the state variables in the set $\{v_2, v_3, v_5, v_7, v_8\}$.

\begin{table}
\begin{center}
\begin{tabular}{clcc}
\vspace{2pt}
{\sc Variable} & {\sc Operators} & $V_0^{v_i}$ & $V_1^{v_i}$\\
\hline
\vspace{2pt}
$v_1$ & $a_1^{v_1} = \langle (v_1=0); (v_1=1) \rangle$ & $V$ & $V$\\
\vspace{2pt}
 & $a_0^{v_1} = \langle (v_1=1); (v_1=0) \rangle$ & & \\
\vspace{2pt}
$v_2$ & $a_1^{v_2} = \langle (v_1=1, v_2=0); (v_2=1) \rangle$ &
 $\emptyset$ & $V$\\
\vspace{2pt}
$v_3$ & $a_1^{v_3} = \langle (v_1=0, v_2=1, v_3=0); (v_3=1) \rangle$ &
 $\{v_4, v_5\}$ & $\{v_6, v_7, v_8\}$\\
\vspace{2pt}
$v_4$ & & $V - \{v_4\}$ & $\emptyset$\\
\vspace{2pt}
$v_5$ & $a_1^{v_5} = \langle (v_3=0, v_4=0, v_5=0); (v_5=1) \rangle$ &
 $\emptyset$ & $\emptyset$\\
\vspace{2pt}
$v_6$ & $a_1^{v_6} = \langle (v_3=1, v_6=0); (v_6=1) \rangle$ &
 $V$ & $V$\\
\vspace{2pt}
 & $a_0^{v_6} = \langle (v_3=1, v_6=1); (v_6=0) \rangle$ & & \\
\vspace{2pt}
$v_7$ & $a_1^{v_7} = \langle (v_6=1, v_7=0); (v_7=1) \rangle$ &
 $\emptyset$ & $V$\\
\vspace{2pt}
$v_8$ & $a_1^{v_8} = \langle (v_6=0, v_7=1, v_8=0); (v_8=1) \rangle$ &
 $\emptyset$ & $\emptyset$\\
\end{tabular}
\caption{Operators and the sets $V_0^{v_i}$ and $V_1^{v_i}$ for the
example planning problem.}
\label{table:ex}
\end{center}
\end{table}

\section{Plan Generation for \THREES}

In this section, we present a polynomial-time algorithm for plan
generation in \THREES. The algorithm produces a solution to any
instance of \THREES{} in the form of a system of macros. The idea is
to construct unary macros that each change the value of a single state
variable. The macros may change the values of other state variables
during execution, but always reset them before terminating. Once the
macros have been generated, the goal can be achieved one state
variable at a time. We show that the algorithm generates a valid plan
if and only if one exists.

We begin by defining macros as we use them in the paper. Next, we
describe the algorithm in pseudo-code (Figures \ref{fig:macro},
\ref{fig:macro3s}, and \ref{fig:finalplan}) and prove its correctness.
To facilitate reading we have moved a straightforward but involving
proof to the appendix.  Following the description of the algorithm we
analyze the complexity of all steps involved. In what follows, we
assume that \THREES{} planning problems are in normal form as defined
in the previous section.

\subsection{Macros}

A macro-operator, or macro for short, is an ordered sequence of
operators viewed as a unit. Each operator in the sequence has to
respect the pre-conditions of operators that follow it, so that no
pre-condition of any operator in the sequence is violated. Applying a
macro is equivalent to applying all operators in the sequence in the
given order. Semantically, a macro is equivalent to a standard
operator in that it has a pre-condition and a post-condition,
unambiguously induced by the pre- and post-conditions of the operators
in its sequence.

Since macros are functionally operators, the operator sequence
associated with a macro can include other macros, as long as this does
not create a circular definition. Consequently, it is possible to
create hierarchies of macros in which the operator sequences of macros
on one level include macros on the level below. The solution to a
planning problem can itself be viewed as a macro which sits at the top
of the hierarchy.

To define macros we first introduce the concept of induced pre- and
post-conditions of operator sequences. If $\Pi=\langle a_1, \ldots,
a_k\rangle$ is an operator sequence, we write $\Pi_i$, $1 \leq i \leq
k$, to denote the subsequence $\langle a_1, \ldots, a_{i}\rangle$.

\begin{definition}
An operator sequence $\Pi=\langle a_1, \ldots, a_k \rangle$ induces a
pre-condition $pre(\Pi) = pre(a_k) \oplus \cdots \oplus pre(a_1)$ and
a post-condition $post(\Pi) = post(a_1) \oplus \cdots \oplus
post(a_k)$. In addition, the operator sequence is well-defined if and
only if $(pre(\Pi_{i-1})\oplus post(\Pi_{i-1})) \triangledown
pre(a_i)$ for each $1 < i \leq k$.
\end{definition}

In what follows, we assume that $P=(V, init, goal, A)$ is a planning
problem such that $V_{post(a)}\subseteq V_{pre(a)}$ for each operator
$a\in A$, and that $\Pi = \langle a_1, \ldots, a_k \rangle$ is an
operator sequence.

\begin{lemma}
\label{lemma:subset}
For each planning problem $P$ of this type and each $\Pi$,
$V_{post(\Pi)} \subseteq V_{pre(\Pi)}$.
\end{lemma}

\proof A direct consequence of the definitions $V_{pre(\Pi)} =
V_{pre(a_1)} \cup \cdots \cup V_{pre(a_k)}$ and $V_{post(\Pi)} =
V_{post(a_1)} \cup \cdots \cup V_{post(a_k)}$.  \endproof

\begin{lemma}
\label{lemma:applicable}
The operator sequence $\Pi$ is applicable in state $s$ if and only if
$\Pi$ is well-defined and $s \triangledown pre(\Pi)$. The state $s_k$
resulting from the application of $\Pi$ to $s$ is $s_k=s \oplus
post(\Pi)$.
\end{lemma}

\proof By induction on $k$. The result clearly holds for $k=1$.  For
$k>1$, note that $pre(\Pi)=pre(a_k)\oplus pre(\Pi_{k-1})$,
$post(\Pi)=post(\Pi_{k-1}) \oplus post(a_k)$, and $\Pi$ is
well-defined if and only if $\Pi_{k-1}$ is well-defined and
$(pre(\Pi_{k-1})\oplus post(\Pi_{k-1}))\triangledown pre(a_k)$.

By hypothesis of induction the state $s_{k-1}$ resulting from the
application of $\Pi_{k-1}$ to $s$ is $s_{k-1}=s \oplus
post(\Pi_{k-1})$. It follows that $s_k = s_{k-1} \oplus post(a_k) = s
\oplus post(\Pi)$.

Assume $\Pi$ is applicable in state $s$. This means that $\Pi_{k-1}$
is applicable in $s$ and that $a_k$ is applicable in $s_{k-1} = s
\oplus post(\Pi_{k-1})$. By hypothesis of induction, the former implies
that $s\triangledown pre(\Pi_{k-1})$ and $\Pi_{k-1}$ is well-defined,
and the latter that $(s \oplus post(\Pi_{k-1})) \triangledown
pre(a_k)$.  This last condition implies that $(pre(\Pi_{k-1}) \oplus
post(\Pi_{k-1})) \triangledown pre(a_k)$ if we use that
$pre(\Pi_{k-1})\sqsubseteq s$, which is a consequence of
$s\triangledown pre(\Pi_{k-1})$ and $s$ being a total state. Finally,
we deduce $s \triangledown (pre(a_k) \oplus pre(\Pi_{k-1}))$ from
$s\triangledown pre(\Pi_{k-1})$ and $(s \oplus post(\Pi_{k-1}))
\triangledown pre(a_k)$, by using that $V_{post(\Pi_{k-1})}\subseteq
V_{pre(\Pi_{k-1})}$. It follows that $\Pi$ is well-defined and that
$s\triangledown pre(\Pi)$.

Conversely, assume that $\Pi$ is well-defined and $s\triangledown
pre(\Pi)$. This implies that $\Pi_{k-1}$ is well-defined and
$s\triangledown pre(\Pi_{k-1})$, so by hypothesis of induction,
$\Pi_{k-1}$ is applicable in state $s$. It remains to
show that $a_k$ is applicable in state $s_{k-1}$, that is, $(s\oplus
post(\Pi_{k-1})) \triangledown pre(a_k)$. From $(pre(\Pi_{k-1})\oplus
post(\Pi_{k-1}))\triangledown pre(a_k)$ it follows that
$post(\Pi_{k-1})\triangledown pre(a_k)$. The fact that
$s\triangledown (pre(a_k)\oplus pre(\Pi_{k-1}))$ and
$V_{post(\Pi_{k-1})}\subseteq V_{pre(\Pi_{k-1})}$ completes the proof.

\endproof

Since macros have induced pre- and post-conditions, Lemmas
\ref{lemma:subset} and \ref{lemma:applicable} trivially extend to the
case for which the operator sequence $\Pi$ includes macros. Now
we are ready to introduce our definition of macros:

\begin{definition}
\label{def:macro}
A macro $m$ is a sequence $\Pi = \langle a_1, \ldots, a_k \rangle$ of
operators and other macros that induces a pre-condition $pre(m) =
pre(\Pi)$ and a post-condition $post(m) = post(\Pi) - pre(\Pi)$. The
macro is well-defined if and only if no circular definitions occur and
$\Pi$ is well-defined.
\end{definition}

To make macros consistent with standard operators, the induced
post-condition should only include state variables whose values are
indeed changed by the macro, which is achieved by computing the
difference between $post(\Pi)$ and $pre(\Pi)$. In particular, it holds
that for a \THREES{} planning problem in normal form, derived macros
satisfy the second condition of normal form, namely that
$post(m)=(v=x)$, $x \in \{0,1\}$, implies $pre(m)(v)=1-x$.

\begin{definition}
Let $Anc^v$ be the set of ancestors of a state variable $v$ in a
\THREES{} planning problem. We define the partial state $pre^v$ on
$V_{pre^v}=Anc^v$ as
\begin{enumerate}
\item $pre^v(u)=1$ if $u \in Anc^v$ is splitting and $v\in V^u_1$,
\item $pre^v(u)=0$ otherwise.
\end{enumerate}
\end{definition}

\begin{definition}
A macro $m$ is a {\em \THREES-macro} if it is well-defined and, for $x
\in \{0,1\}$, $post(m)=(v=x)$ and $pre(m)\sqsubseteq
pre^v\oplus(v=1-x)$.
\end{definition}

The algorithm we present only generates \THREES-macros. In fact, it
generates at most one macro $m=m^v_x$ with $post(m)=(v=x)$ for each
state variable $v$ and value $x\in \{0, 1\}$. To illustrate the idea
of \THREES-macros and give a flavor of the algorithm, Table
\ref{table:macros} lists the macros generated by the algorithm in the
example \THREES{} planning problem from the previous section.

\begin{table}
\begin{center}
\begin{tabular}{cllc}
\vspace{4pt}
{\sc Macro} & {\sc Sequence} & {\sc Pre-condition} & {\sc Post-condition}\\
\hline
\vspace{2pt}
$m_1^{v_1}$ & $\langle a_1^{v_1} \rangle$ & $(v_1=0)$ & $(v_1=1)$\\
\vspace{2pt}
$m_0^{v_1}$ & $\langle a_0^{v_1} \rangle$ & $(v_1=1)$ & $(v_1=0)$\\
\vspace{2pt}
$m_1^{v_2}$ & $\langle m_1^{v_1}, a_1^{v_2}, m_0^{v_1} \rangle$ &
 $(v_1=0, v_2=0)$ & $(v_2=1)$\\
\vspace{2pt}
$m_1^{v_3}$ & $\langle a_1^{v_3} \rangle$ &
 $(v_1=0, v_2=1, v_3=0)$ & $(v_3=1)$\\
\vspace{2pt}
$m_1^{v_5}$ & $\langle a_1^{v_5} \rangle$ &
 $(v_3=0, v_4=0, v_5=0)$ & $(v_5=1)$\\
\vspace{2pt}
$m_1^{v_6}$ & $\langle a_1^{v_6} \rangle$ &
 $(v_3=1, v_6=0)$ & $(v_6=1)$\\
\vspace{2pt}
$m_0^{v_6}$ & $\langle a_0^{v_6} \rangle$ &
 $(v_3=1, v_6=1)$ & $(v_6=0)$\\
\vspace{2pt}
$m_1^{v_7}$ & $\langle m_1^{v_6}, a_1^{v_7}, m_0^{v_6} \rangle$ &
 $(v_3=1, v_6=0, v_7=0)$ & $(v_7=1)$\\
\vspace{2pt}
$m_1^{v_8}$ & $\langle a_1^{v_8} \rangle$ &
 $(v_3=1, v_6=0, v_7=1, v_8=0)$ &
 $(v_8=1)$\\
\end{tabular}
\caption{Macros generated by the algorithm in the example planning problem.}
\label{table:macros}
\end{center}
\end{table}

We claim that each macro is a \THREES-macro. For example, the operator
sequence $\langle a_1^{v_6} \rangle$ induces a pre-condition $(v_3=1,
v_6=0)$ and a post-condition $(v_3=1, v_6=0) \oplus (v_6=1) =
(v_3=1,v_6=1)$. Thus, the macro $m_1^{v_6}$ induces a pre-condition
$pre(m_1^{v_6}) = (v_3=1, v_6=0)$ and a post-condition
$post(m_1^{v_6}) = (v_3=1,v_6=1) - (v_3=1, v_6=0) = (v_6=1)$. Since
$v_2$ and $v_3$ are splitting and since $v_6 \in V_1^{v_2}$ and
$v_6 \in V_1^{v_3}$, it follows that $pre^{v_6} \oplus (v_6=0) =
(v_1=0, v_2=1, v_3=1, v_6=0)$, so $pre(m_1^{v_6}) = (v_3=1, v_6=0)
\sqsubseteq pre^{v_6} \oplus (v_6=0)$.

The macros can be combined to produce a solution to the planning
problem. The idea is to identify each state variable $v$ such that
$goal(v) = 1$ and append the macro $m_1^v$ to the solution plan. In
the example, this results in the operator sequence $\langle m_1^{v_5},
m_1^{v_8} \rangle$. However, the pre-condition of $m_1^{v_8}$
specifies $v_3=1$ and $v_7=1$, which makes it necessary to insert
$m_1^{v_3}$ and $m_1^{v_7}$ before $m_1^{v_8}$. In addition, the
pre-condition of $m_1^{v_3}$ specifies $v_2=1$, which makes it
necessary to insert $m_1^{v_2}$ before $m_1^{v_3}$, resulting in the
final plan $\langle m_1^{v_5}, m_1^{v_2}, m_1^{v_3}, m_1^{v_7},
m_1^{v_8} \rangle$. Note that the order of the macros matter;
$m_1^{v_5}$ requires $v_3$ to be 0 while $m_1^{v_8}$ requires $v_3$ to
be 1. For a splitting state variable $v$, the goal state should be
achieved for state variables in $V_0^v$ before the value of $v$ is set
to 1. We can expand the solution plan so that it consists solely of
operators in $A$. In our example, this results in the operator
sequence $\langle a_1^{v_5}, a_1^{v_1}, a_1^{v_2}, a_0^{v_1},
a_1^{v_3}, a_1^{v_6}, a_1^{v_7}, a_0^{v_6}, a_1^{v_8} \rangle$. In
this case, the algorithm generates an optimal plan, although this is
not true in general.

\subsection{Description of the Algorithm}

We proceed by providing a detailed description of the algorithm for
plan generation in \THREES. We first describe the subroutine for
generating a unary macro that sets the value of a state variable $v$
to $x$. This algorithm, which we call {\sc GenerateMacro}, is
described in Figure \ref{fig:macro}. The algorithm takes as input a
planning problem $P$, a state variable $v$, a value $x$ (either 0 or
1), and a set of macros $M$ for $v$'s ancestors in the causal graph.
Prior to executing the algorithm, we perform a topological sort of the
state variables. We assume that, for each $v \in V$ and $x \in
\{0,1\}$, $M$ contains at most one macro $m_x^v$ such that
$post(m_x^v) = (v = x)$. In the algorithm, we use the notation $m_x^v
\in M$ to test whether or not $M$ contains $m_x^v$.

\begin{figure}
\begin{tabular}{ll}
1 & {\bf function} {\sc GenerateMacro}($P$, $v$, $x$, $M$)\\
2 & {\bf for each} $a \in A$ such that $post(a)(v) = x$ {\bf do}\\
3 & \hspace{15pt} $S_0 \leftarrow S_1 \leftarrow \langle \rangle$\\
4 & \hspace{15pt} $satisfy \leftarrow$ true\\
5 & \hspace{15pt} $U \leftarrow
 \{u \in V_{pre(a)} - \{v\} \mid pre(a)(u) = 1\}$\\
6 & \hspace{15pt} {\bf for each} $u \in U$ in increasing topological order
 {\bf do}\\
7 & \hspace{30pt} {\bf if} $u$ is static {\bf or} $m_1^u \notin M$
 {\bf then}\\
8 & \hspace{45pt} $satisfy \leftarrow$ false\\
9 & \hspace{30pt} {\bf else if} $u$ is not splitting {\bf and} $m_0^u \in M$ 
 {\bf and} $m_1^u \in M$ {\bf then}\\
10 & \hspace{45pt} $S_0 \leftarrow \langle S_0, m_0^u \rangle$\\
11 & \hspace{45pt} $S_1 \leftarrow \langle m_1^u, S_1 \rangle$\\
12 & \hspace{15pt} {\bf if} $satisfy$ {\bf then}\\
13 & \hspace{30pt} {\bf return} $\langle S_1, a, S_0 \rangle$\\
14 & {\bf return} $fail$\\
\end{tabular}
\caption{Algorithm for generating a macro that sets the value of $v$ to $x$.}
\label{fig:macro}
\end{figure}

For each operator $a \in A$ that sets the value of $v$ to $x$, the
algorithm determines whether it is possible to satisfy its
pre-condition $pre(a)$ starting from the initial state. To do this,
the algorithm finds the set $U$ of state variables to which $pre(a)$
assigns 1 (the values of all other state variables already satisfy
$pre(a)$ in the initial state). The algorithm constructs two sequences
of operators, $S_0$ and $S_1$, by going through the state variables of
$U$ in increasing topological order.  If $S$ is an operator sequence,
we use $\langle S, o \rangle$ as shorthand to denote an operator
sequence of length $|S| + 1$ consisting of all operators of $S$
followed by $o$, which can be either an operator or a macro. If it is
possible to satisfy the pre-condition $pre(a)$ of some operator $a \in
A$, the algorithm returns the macro $\langle S_1, a, S_0
\rangle$. Otherwise, it returns $fail$.

\begin{lemma}
\label{lemma:symm}
If $v$ is symmetrically reversible and {\sc GenerateMacro}($P$, $v$,
1, $M$) successfully generates a macro, so does {\sc
GenerateMacro}($P$, $v$, 0, $M$).
\end{lemma}

\proof
Assume that {\sc GenerateMacro}($P$, $v$, 1, $M$) successfully returns
the macro $\langle S_1, a, S_0 \rangle$ for some operator $a \in A$
such that $post(a) = 1$. From the definition of symmetrically
reversible it follows that there exists an operator $a' \in A$ such
that $post(a') = 0$ and $pre(a') \mid V - \{v\} = pre(a) \mid V -
\{v\}$. Thus, the set $U$ is identical for $a$ and $a'$. As a
consequence, the values of $S_0$, $S_1$, and $satisfy$ are the same
after the loop, which means that {\sc GenerateMacro}($P$, $v$, 0, $M$)
returns the macro $\langle S_1, a', S_0 \rangle$ for $a'$. Note that
{\sc GenerateMacro}($P$, $v$, 0, $M$) may return another macro if it
goes through the operators of $A$ in a different order; however, it is
guaranteed to successfully return a macro.
\endproof

\begin{theorem}
\label{thm:3s}
If the macros in $M$ are \THREES-macros and {\sc GenerateMacro}($P$,
$v$, $x$, $M$) generates a macro $m^v_x\neq fail$, then $m^v_x$ is a
\THREES-macro.
\end{theorem}

The proof of Theorem \ref{thm:3s} appears in Appendix
\ref{app:thm}.\\

\begin{figure}
\begin{tabular}{ll}
1 & {\bf function} {\sc Macro-\THREES}($P$)\\
2 & $M \leftarrow \emptyset$\\
3 & {\bf for each} $v \in V$ in increasing topological order {\bf do}\\
4 & \hspace{15pt} $m_1^v \leftarrow$ {\sc GenerateMacro}($P$, $v$, 1, $M$)\\
5 & \hspace{15pt} $m_0^v \leftarrow$ {\sc GenerateMacro}($P$, $v$, 0, $M$)\\
6 & \hspace{15pt} {\bf if} $m_1^v \neq fail$ {\bf and} $m_0^v \neq fail$
 {\bf then}\\
7 & \hspace{30pt} $M \leftarrow M \cup \{m_1^v, m_0^v\}$\\
8 & \hspace{15pt} {\bf else if} $m_1^v \neq fail$ {\bf and}
 $goal(v) \neq 0$ {\bf then}\\
9 & \hspace{30pt} $M \leftarrow M \cup \{m_1^v\}$\\
10 & {\bf return} {\sc GeneratePlan}($P$, $V$, $M$)\\
\end{tabular}
\caption{The algorithm {\sc Macro-\THREES}.}
\label{fig:macro3s}
\end{figure}

Next, we describe the algorithm for plan generation in \THREES, which
we call {\sc Macro-\THREES}. Figure \ref{fig:macro3s} shows pseudocode
for {\sc Macro-\THREES}. The algorithm goes through the state
variables in increasing topological order and attempts to generate two
macros for each state variable $v$, $m_1^v$ and $m_0^v$. If both
macros are successfully generated, they are added to the current set
of macros $M$. If only $m_1^v$ is generated and the goal state does
not assign 0 to $v$, the algorithm adds $m_1^v$ to $M$. Finally, the
algorithm generates a plan using the subroutine {\sc GeneratePlan},
which we describe later.

\begin{lemma}
\label{lemma:exists}
Let $P$ be a \THREES{} planning problem and let $v \in V$ be a state
variable. If there exists a valid plan for solving $P$ that sets $v$
to 1, {\sc Macro-\THREES}($P$) adds the macro $m_1^v$ to $M$. If, in
addition, the plan resets $v$ to 0, {\sc Macro-\THREES}($P$) adds
$m_0^v$ to $M$.
\end{lemma}

\proof
First note that if $m_1^v$ and $m_0^v$ are generated, {\sc
Macro-\THREES}($P$) adds them both to $M$. If $m_1^v$ is generated but
not $m_0^v$, {\sc Macro-\THREES}($P$) adds $m_1^v$ to $M$ unless
$goal(v) = 0$. However, $goal(v) = 0$ contradicts the fact that there
is a valid plan for solving $P$ that sets $v$ to 1 without resetting
it to 0. It remains to show that {\sc GenerateMacro}($P$, $v$, 1, $M$)
always generates $m_1^v \neq fail$ and that {\sc GenerateMacro}($P$,
$v$, 0, $M$) always generates $m_0^v \neq fail$ if the plan resets $v$
to 0.

A plan for solving $P$ that sets $v$ to 1 has to contain an operator
$a \in A$ such that $post(a)(v) = 1$. If the plan also resets $v$ to
0, it has to contain an operator $a' \in A$ such that $post(a')(v) =
0$. We show that {\sc GenerateMacro}($P$, $v$, 1, $M$) successfully
generates $m_1^v \neq fail$ if $a$ is the operator selected on line
2. Note that the algorithm may return another macro if it selects
another operator before $a$; however, if it always generates a macro
for $a$, it is guaranteed to successfully return a macro $m_1^v \neq
fail$. The same is true for $m_0^v$ and $a'$.

We prove the lemma by induction on state variables $v$. If $v$ has no
ancestors in the causal graph, the set $U$ is empty by default. Thus,
$satisfy$ is never set to false and {\sc GenerateMacro}($P$, $v$, 1,
$M$) successfully returns the macro $m_1^v = \langle a \rangle$ for
$a$. If $a'$ exists, {\sc GenerateMacro}($P$, $v$, 0, $M$)
successfully returns $m_0^v = \langle a' \rangle$ for $a'$.

If $v$ has ancestors in the causal graph, let $U = \{u \in V_{pre(a)}
- \{v\} \mid pre(a)(u) = 1\}$. Since the plan contains $a$ it has to
set each $u \in U$ to 1. By hypothesis of induction, {\sc
Macro-\THREES}($P$) adds $m_1^u$ to $M$ for each $u \in U$. As a
consequence, $satisfy$ is never set to false and thus, {\sc
GenerateMacro}($P$, $v$, 1, $M$) successfully returns $m_1^v$ for $a$.
If $a'$ exists, let $W = \{w \in V_{pre(a')} - \{v\} \mid pre(a')(w) =
1\}$.  If the plan contains $a'$, it has to set each $w
\in W$ to 1.  By hypothesis of induction, {\sc Macro-\THREES}($P$)
adds $m_1^w$ to $M$ for each $w \in W$ and consequently, {\sc
GenerateMacro}($P$, $v$, 0, $M$) successfully returns $m_0^v$ for
$a'$.\\
\endproof

\begin{figure}
\begin{tabular}{ll}
1 & {\bf function} {\sc GeneratePlan}($P$, $W$, $M$)\\
2 & {\bf if} $|W| = 0$ {\bf then}\\
3 & \hspace{15pt} {\bf return} $\langle \rangle$\\
4 & $v \leftarrow$ first variable in topological order present in $W$\\
5 & {\bf if} $v$ is splitting {\bf then}\\
6 & \hspace{15pt} $\Pi_0^v \leftarrow$
 {\sc Generate-Plan}($P$, $W \cap (V_0^v - \{v\})$, $M$)\\
7 & \hspace{15pt} $\Pi_1^v \leftarrow$
 {\sc Generate-Plan}($P$, $W \cap (V_1^v - \{v\})$, $M$)\\
8 & \hspace{15pt} $\Pi_*^v \leftarrow$
 {\sc Generate-Plan}($P$, $W \cap (V- V_0^v - V_1^v - \{v\})$, $M$)\\
9 & \hspace{15pt} {\bf if} $\Pi_0^v = fail$ {\bf or} $\Pi_1^v = fail$ {\bf or}
 $\Pi_*^v = fail$ {\bf or} ($goal(v) = 1$ {\bf and} $m_1^v \notin M$)
 {\bf then}\\
10 & \hspace{30pt} {\bf return} $fail$\\
11 & \hspace{15pt} {\bf else if} $m_1^v \notin M$ {\bf then return}
 $\langle \Pi_*^v, \Pi_0^v, \Pi_1^v \rangle$\\
12 & \hspace{15pt} {\bf else if} $goal(v) = 0$ {\bf then return}
 $\langle \Pi_*^v, \Pi_0^v, m_1^v, \Pi_1^v, m_0^v \rangle$\\
13 & \hspace{15pt} {\bf else return}
 $\langle \Pi_*^v, \Pi_0^v, m_1^v, \Pi_1^v \rangle$\\
14 & $\Pi \leftarrow$ {\sc Generate-Plan}($P$, $W - \{v\}$, $M$)\\
15 & {\bf if} $\Pi = fail$ {\bf or}
 ($goal(v) = 1$ {\bf and} $m_1^v \notin M$) {\bf then return} $fail$\\
16 & {\bf else if} $goal(v) = 1$ {\bf then return}
 $\langle \Pi, m_1^v \rangle$\\
17 & {\bf else return} $\Pi$\\
\end{tabular}
\caption{Algorithm for generating the final plan}
\label{fig:finalplan}
\end{figure}

Finally, we describe the subroutine {\sc GeneratePlan}($P$, $W$, $M$)
for generating the final plan given a planning problem $P$, a set of
state variables $W$ and a set of macros $M$. If the set of state
variables is empty, {\sc GeneratePlan}($P$, $W$, $M$) returns an empty
operator sequence. Otherwise, it finds the state variable $v \in W$
that comes first in topological order. If $v$ is splitting, the
algorithm separates $W$ into the three sets described by $V_0^v$,
$V_1^v$, and $V_*^v = V - V_0^v - V_1^v$. The algorithm recursively
generates plans for the three sets and if necessary, inserts $m_1^v$
between $V_0^v$ and $V_1^v$ in the final plan. If this is not the
case, the algorithm recursively generates a plan for $W - \{v\}$. If
$goal(v) = 1$ and $m_1^v$, the algorithm appends $m_1^v$ to the end of
the resulting plan.

\begin{lemma}
\label{lemma:prepost}
Let $\Pi^W$ be the plan generated by {\sc GeneratePlan}($P$, $W$,
$M$), let $v$ be the first state variable in topological order present
in $W$, and let $\Pi^V = \langle \Pi^a, \Pi^W, \Pi^b \rangle$ be the
final plan generated by {\sc Macro-\THREES}($P$). If $m_1^v \in M$ it
follows that $(pre(\Pi^a) \oplus post(\Pi^a)) \triangledown
pre(m_1^v)$.
\end{lemma}

\proof
We determine the content of the operator sequence $\Pi^a$ that
precedes $\Pi^W$ in the final plan by inspection. Note that the call
{\sc GeneratePlan}($P$, $W$, $M$) has to be nested within a sequence
of recursive calls to {\sc GeneratePlan} starting with {\sc
GeneratePlan}($P$, $V$, $M$).\linebreak Let $Z$ be the set of state
variables such that each $u \in Z$ was the first state variable in
topological order for some call to {\sc GeneratePlan} prior to {\sc
GeneratePlan}($P$, $W$, $M$). Each $u \in Z$ has to correspond to a
call to {\sc GeneratePlan} with some set of state variables $U$ such
that $W \subset U$. If $u$ is not splitting, $u$ does not contribute
to $\Pi^a$ since the only possible addition of a macro to the plan on
line 16 places the macro $m_1^u$ at the end of the plan generated
recursively.

Assume that $u \in Z$ is a splitting state variable. We have three
cases: $W \subseteq V_0^u$, $W \subseteq V_1^u$, and $W \subseteq
V_*^u = V - V_0^u - V_1^u$. If $W \subseteq V_*^u$, $u$ does not
contribute to $\Pi^a$ since it never places macros before $\Pi_*^u$.
If $W \subseteq V_0^u$, the plan $\Pi_*^u$ is part of $\Pi^a$ since it
precedes $\Pi_0^u$ on lines 11, 12, and 13. If $W \subseteq V_1^u$,
the plans $\Pi_*^u$ and $\Pi_0^u$ are part of $\Pi^a$ since they both
precede $\Pi_1^u$ in all cases. If $m_1^u \in M$, the macro $m_1^u$ is
also part of $\Pi^a$ since it precedes $\Pi_1^u$ on lines 12 and 13.
No other macros are part of $\Pi^a$.

Since the macros in $M$ are unary, the plan generated by {\sc
GeneratePlan}($P$, $U$, $M$) only changes the values of state
variables in $U$. For a splitting state variable $u$, there are no
edges from $V_*^u - \{u\}$ to $V_0^u$, from $V_*^u - \{u\}$ to
$V_1^u$, or from $V_0^u$ to $V_1^u$. It follows that the plan
$\Pi_*^u$ does not change the value of any state variable that appears
in the pre-condition of a macro in $\Pi_0^u$. The same holds for
$\Pi_*^u$ with respect to $\Pi_1^u$ and for $\Pi_0^u$ with respect to
$\Pi_1^u$. Thus, the only macro in $\Pi^a$ that changes the value of a
splitting state variable $u \in Anc^v$ is $m_1^u$ in case $W \subseteq
V_1^u$.

Recall that $pre^v$ is defined on $Anc^v$ and assigns 1 to $u$ if and
only if $u$ is splitting and $v \in V_1^u$. For all other ancestors of
$v$, the value 0 holds in the initial state and is not altered by
$\Pi^a$. If $u$ is splitting and $v \in V_1^u$, it follows from the
definition of \THREES-macros that $pre(m_1^v)(u) = 1$ or
$pre(m_1^v)(u) = \perp$. If $pre(m_1^v)(u) = 1$, it is correct to
append $m_1^u$ before $m_1^v$ to satisfy $pre(m_1^v)(u)$. If $m_1^u
\notin M$ it follows that $u \notin V_{pre(m_1^v)}$, since
$pre(m_1^v)(u) = 1$ would have caused {\sc GenerateMacro}($P$, $v$,
$1$, $M$) to set $satisfy$ to false on line 8. Thus, the pre-condition
$pre(m_1^v)$ of $m_1^v$ agrees with $pre(\Pi^a) \oplus post(\Pi^a)$ on
the value of each state variable, which means that the two partial
states match.
\endproof

\begin{lemma}
\label{lemma:welldef}
{\sc GeneratePlan}($P$, $V$, $M$) generates a well-defined plan.
\end{lemma}

\proof
Note that for each state variable $v \in V$, {\sc GeneratePlan}($P$,
$W$, $M$) is called precisely once such that $v$ is the first state
variable in topological order. From Lemma \ref{lemma:prepost} it
follows that $(pre(\Pi^a) \oplus post(\Pi^a)) \triangledown
pre(m_1^v)$, where $\Pi^a$ is the plan that precedes $\Pi^W$ in the
final plan. Since $v$ is the first state variable in topological order
in $W$, the plans $\Pi_0^v$, $\Pi_1^v$, $\Pi_*^v$, and $\Pi$,
recursively generated by {\sc GeneratePlan}, do not change the value
of any state variable in $pre(m_1^v)$. It follows that $m_1^v$ is
applicable following $\langle \Pi^a, \Pi_*^v, \Pi_0^v \rangle$ or
$\langle \Pi^a, \Pi \rangle$. Since $m_1^v$ only changes the value of
$v$, $m_0^v$ is applicable following $\langle \Pi^a, \Pi_*^v, \Pi_0^v,
m_1^v, \Pi_1^v \rangle$.
\endproof

\begin{theorem}
{\sc Macro-\THREES}($P$) generates a valid plan for solving a planning
problem in \THREES{} if and only if one exists.
\end{theorem}

\proof
{\sc GeneratePlan}($P$, $V$, $M$) returns $fail$ if and only if there
exists a state variable $v \in V$ such that $goal(v) = 1$ and $m_1^v
\notin M$. From Lemma \ref{lemma:exists} it follows that there does
not exist a valid plan for solving $P$ that sets $v$ to 1.
Consequently, there does not exist a plan for solving $P$. Otherwise,
{\sc GeneratePlan}($P$, $V$, $M$) returns a well-defined plan due to
Lemma \ref{lemma:welldef}. Since the plan sets to 1 each state
variable $v$ such that $goal(v) = 1$ and resets to 0 each state
variable $v$ such that $goal(v) = 0$, the plan is a valid plan for
solving the planning problem.
\endproof

\subsection{Examples}
\label{sec:examples}

\begin{figure}
\centerline{\epsfig{figure=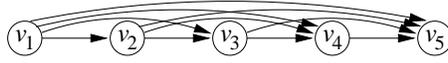,height=0.75cm}}
\vspace{16pt}
\caption{Causal graph of the planning problem $P_5$.}
\label{fig:chain}
\end{figure}

We illustrate the algorithm on an example introduced by
\citeA{Jonsson98b} to show that there are instances of \THREES{} with
exponentially sized minimal solutions.  Let $P_n = \langle V, init,
goal, A \rangle$ be a planning problem defined by a natural number
$n$, $V = \{v_1, \ldots, v_n\}$, and a goal state defined by $V_{goal}
= V$, $goal(v_i) = 0$ for each $v_i \in \{v_1, \ldots, v_{n-1}\}$, and
$goal(v_n) = 1$.  For each state variable $v_i \in V$, there are two
operators in $A$:

\[
a_1^{v_i} = \langle (v_1 = 0, \ldots, v_{i-2} = 0, v_{i-1} = 1, v_i = 0);
                    (v_i = 1) \rangle,
\]
\[
a_0^{v_i} = \langle (v_1 = 0, \ldots, v_{i-2} = 0, v_{i-1} = 1, v_i = 1);
                    (v_i = 0) \rangle.
\]

\noindent
In other words, each state variable is symmetrically reversible. The
causal graph of the planning problem $P_5$ is shown in Figure
\ref{fig:chain}. Note that for each state variable $v_i \in \{v_1,
\ldots, v_{n-2}\}$, $pre(a_1^{v_{i+1}})(v_i) = 1$ and
$pre(a_1^{v_{i+2}})(v_i) = 0$, so $v_{i+1} \in Q_1^{v_i}$ and $v_{i+2}
\in Q_0^{v_i}$. Since there is an edge in the causal graph between
$v_{i+1}$ and $v_{i+2}$, no state variable in $\{v_1, \ldots,
v_{n-2}\}$ is splitting. On the other hand, $v_{n-1}$ and $v_n$ are
splitting since $V_0^{v_{n-1}} = \emptyset$ and $V_0^{v_n} = V_1^{v_n}
= \emptyset$. \citeA{Backstrom95} showed that the length of the
shortest plan solving $P_n$ is $2^n - 1$, i.e., exponential in the
number of state variables.

For each state variable $v_i \in \{v_1, \ldots, v_{n-1}\}$, our
algorithm generates two macros $m_1^{v_i}$ and $m_0^{v_i}$. There is a
single operator, $a_1^{v_i}$, that changes the value of $v_i$ from 0
to 1.  $pre(a_1^{v_i})$ only assigns 1 to $v_{i-1}$, so $U =
\{v_{i-1}\}$. Since $v_{i-1}$ is not splitting, $m_1^{v_i}$ is defined
as $m_1^{v_i} = \langle m_1^{v_{i-1}}, a_1^{v_i}, m_0^{v_{i-1}}
\rangle$. Similarly, $m_0^{v_i}$ is defined as $m_0^{v_i} = \langle
m_1^{v_{i-1}}, a_0^{v_i}, m_0^{v_{i-1}} \rangle$. For state variable
$v_n$, $U = \{v_{n-1}\}$, which is splitting, so $m_1^{v_n}$ is
defined as $m_1^{v_n} = \langle a_1^{v_n} \rangle$.

To generate the final plan, the algorithm goes through the state
variables in topological order. For state variables $v_1$ through
$v_{n-2}$, the algorithm does nothing, since these state variables are
not splitting and their goal state is not 1. For state variable
$v_{n-1}$, the algorithm recursively generates the plan for $v_n$,
which is $\langle m_1^{v_n} \rangle$ since $goal(v_n) = 1$. Since
$goal(v_{n-1}) = 0$, the algorithm inserts $m_1^{v_{n-1}}$ before
$m_1^{v_n}$ to satisfy its pre-condition $v_{n-1} = 1$ and
$m_0^{v_{n-1}}$ after $m_1^{v_n}$ to achieve the goal state
$goal(v_{n-1}) = 0$. Thus, the final plan is $\langle m_1^{v_{n-1}},
m_1^{v_n}, m_0^{v_{n-1}} \rangle$. If we expand the plan, we end up
with a sequence of $2^n - 1$ operators.  However, no individual macro
has operator sequence length greater than 3.  Together, the macros
recursively specify a complete solution to the planning problem.

We also demonstrate that there are planning problems in \THREES{} with
polynomial length solutions for which our algorithm may generate
exponential length solutions.  To do this, we modify the planning
problem $P_n$ by letting $goal(v_i) = 1$ for each $v_i \in V$. In
addition, for each state variable $v_i \in V$, we add two operators to
$A$:

\[
b_1^{v_i} = \langle (v_1 = 1, \ldots, v_{i-1} = 1, v_i = 0);
                    (v_i = 1) \rangle,
\]
\[
b_0^{v_i} = \langle (v_1 = 1, \ldots, v_{i-1} = 1, v_i = 1);
	            (v_i = 0) \rangle.
\]

We also add an operator $c_1^{v_n} = \langle (v_{n-1} = 0, v_n = 0);
(v_n = 1) \rangle$ to $A$. As a consequence, state variables in
$\{v_1, \ldots, v_{n-2}\}$ are still symmetrically reversible but not
splitting. $v_{n-1}$ is also symmetrically reversible but no longer
splitting, since $pre(a_1^{v_n})(v_{n-1}) = 1$ and
$pre(c_1^{v_n})(v_{n-1}) = 0$ implies that $v_n \in V_0^{v_{n-1}}
\cap V_1^{v_{n-1}}$.  $v_n$ is still splitting since $V_0^{v_n} =
V_1^{v_n} = \emptyset$.  Assume that {\sc GenerateMacro}($P$, $v_i$,
$x$, $M$) always selects $b_x^{v_i}$ first. As a consequence, for each
state variable $v_i \in V$ and each $x \in \{0,1\}$, {\sc
GenerateMacro}($P$, $v_i$, $x$, $M$) generates the macro $m_x^{v_i} =
\langle m_1^{v_{i-1}}, \ldots, m_1^{v_1}, b_x^{v_i}, m_0^{v_1},
\ldots, m_0^{v_{i-1}} \rangle$.

Let $L_i$ be the length of the plan represented by $m_x^{v_i}$, $x \in
\{0,1\}$.  From the definition of $m_x^{v_i}$ above we have that $L_i
= 2(L_1 + \ldots + L_{i-1}) + 1$.  We show by induction that $L_i =
3^{i-1}$.  The length of any macro for $v_1$ is $L_1 = 1 = 3^0$. For
$i > 1$,

\[
L_i = 2(3^0 + \ldots + 3^{i-2}) + 1 =
2\frac{3^{i-1} - 1}{3 - 1} + 1 =
2\frac{3^{i-1} - 1}{2} + 1 = 3^{i-1} - 1 + 1 = 3^{i-1}.
\]

\noindent
To generate the final plan the algorithm has to change the value of
each state variable from 0 to 1, so the total length of the plan is $L
= L_1 + \ldots + L_n = 3^0 + \ldots + 3^{n-1} = (3^n - 1)/2$.
However, there exists a plan of length $n$ that solves the planning
problem, namely $\langle b_1^{v_1}, \ldots, b_1^{v_n} \rangle$.

\subsection{Complexity}

In this section we prove that the complexity of our algorithm is
polynomial. To do this we analyze each step of the algorithm
separately. A summary of the complexity result for each step of the
algorithm is given below. Note that the number of edges $|E|$ in the
causal graph is $O(|A||V|)$, since each operator may introduce
$O(|V|)$ edges. The complexity result $O(|V| + |E|) = O(|A||V|)$ for
topological sort follows from \citeA{Cormen90}.\\

\begin{tabular}{lll}
Constructing the causal graph $G = (V, E)$ & $O(|A||V|)$ &
 Lemma \ref{lemma:cg}\\
Calculating $V_1^v$ and $V_0^v$ for each $v \in V$ & $O(|A||V|^2)$ &
 Lemma \ref{lemma:split}\\
Performing a topological sort of $G$ & $O(|A||V|)$ & \\
{\sc GenerateMacro}($P$, $v$, $x$, $M$) & $O(|A||V|)$ &
 Lemma \ref{lemma:gm}\\
{\sc GeneratePlan}($P$, $V$, $M$) & $O(|V|^2)$ & Lemma \ref{lemma:gp}\\
{\sc Macro-\THREES}($P$) & $O(|A||V|^2)$ & Theorem \ref{thm:macro3s}\\
\end{tabular}

\begin{lemma}
\label{lemma:cg}
The complexity of constructing the causal graph $G = (V, E)$ is
$O(|A||V|)$.
\end{lemma}

\proof
The causal graph consists of $|V|$ nodes. For each operator $a \in A$
and each state variable $u \in V_{pre(a)}$, we should add an edge from
$u$ to the unique state variable $v \in V_{post(a)}$. In the worst
case, $|V_{pre(a)}| = O(|V|)$, in which case the complexity is
$O(|A||V|)$.
\endproof

\begin{lemma}
\label{lemma:split}
The complexity of calculating the sets $V_0^v$ and $V_1^v$ for each
state variable $v \in V$ is $O(|A||V|^2)$.
\end{lemma}

\proof
For each state variable $v \in V$, we have to establish the sets
$Q_0^v$ and $Q_1^v$, which requires going through each operator $a \in
A$ in the worst case. Note that we are only interested in the
pre-condition on $v$ and the unique state variable in $V_{post(a)}$,
which means that we do not need to go through each state variable in
$V_{pre(a)}$. Next, we have to construct the graph $G_0^v$. We can do
this by copying the causal graph $G$, which takes time $O(|A||V|)$,
and removing the edges between $v$ and $Q_0^v - Q_1^v$, which takes
time $O(|V|)$.

Finally, to construct the set $V_0^v$ we should find each state
variable that is weakly connected to some state variable $u \in Q_0^v$
in the graph $G_0^v$. For each state variable $u \in Q_0^v$,
performing an undirected search starting at $u$ takes time
$O(|A||V|)$. Once we have performed search starting at $u$, we only
need to search from state variables in $Q_0^v$ that were not reached
during the search. This way, the total complexity of the search does
not exceed $O(|A||V|)$. The case for constructing $V_1^v$ is
identical.  Since we have to perform the same procedure for each state
variable $v \in V$, the total complexity of this step is
$O(|A||V|^2)$.
\endproof

\begin{lemma}
\label{lemma:gm}
The complexity of {\sc GenerateMacro}($P$, $v$, $x$, $M$) is
$O(|A||V|)$.
\end{lemma}

\proof
For each operator $a \in A$, {\sc GenerateMacro}($P$, $v$, $x$, $M$)
needs to check whether $post(a)(v) = x$. In the worst case, $|U| =
O(|V|)$, in which case the complexity of the algorithm is
$O(|A||V|)$.
\endproof

\begin{lemma}
\label{lemma:gp}
The complexity of {\sc GeneratePlan}($P$, $V$, $M$) is $O(|V|^2)$.
\end{lemma}

\proof
Note that for each state variable $v \in V$, {\sc GeneratePlan}($P$,
$V$, $M$) is called recursively exactly once such that $v$ is the
first variable in topological order. In other words, {\sc
GeneratePlan}($P$, $V$, $M$) is called exactly $|V|$ times. {\sc
GeneratePlan}($P$, $V$, $M$) contains only constant operations except
the intersection and difference between sets on lines 6-8. Since
intersection and set difference can be done in time $O(|V|)$, the
total complexity of {\sc GeneratePlan}($P$, $V$, $M$) is $O(|V|^2)$.
\endproof

\begin{theorem}
\label{thm:macro3s}
The complexity of {\sc Macro-\THREES}($P$) is $O(|A||V|^2)$.
\end{theorem}

\proof
Prior to executing {\sc Macro-\THREES}($P$), it is necessary to
construct the causal graph $G$, find the sets $V_0^v$ and $V_1^v$ for
each state variable $v \in V$, and perform a topological sort of
$G$. We have shown that these steps take time $O(|A||V|^2)$. For each
state variable $v \in V$, {\sc Macro-\THREES}($P$) calls {\sc
GenerateMacro}($P$, $v$, $x$, $M$) twice. From Lemma \ref{lemma:gm} it
follows that this step takes time $O(2|V||A||V|) = O(|A||V|^2)$.
Finally, {\sc Macro-\THREES}($P$) calls {\sc GeneratePlan}($P$, $V$,
$M$), which takes time $O(|V|^2)$ due to Lemma \ref{lemma:gp}. It
follows that the complexity of {\sc Macro-\THREES}($P$) is
$O(|A||V|^2)$.\\
\endproof

We conjecture that it is possible to improve the above complexity
result for {\sc Macro-\THREES}($P$) to $O(|A||V|)$. However, the proof
seems somewhat complex, and our main objective is not to devise an
algorithm that is as efficient as possible. Rather, we are interested
in establishing that our algorithm is polynomial, which follows from
Theorem \ref{thm:macro3s}.

\subsection{Plan Length}

In this section we study the length of the plans generated by the
given algorithm. To begin with, we derive a general bound on the
length of such plans. Then, we show how to compute the actual length
of some particular plan without expanding its macros. We also present
an algorithm that uses this computation to efficiently obtain the
$i$-th action of the plan from its macro form.  We start by
introducing the concept of depth of state variables in the causal
graph.

\begin{definition}
\label{def:depth}
The depth $d(v)$ of a state variable $v$ is the longest path
from $v$ to any other state variable in the causal graph.
\end{definition}

\noindent
Since the causal graph is acyclic for planning problems in \THREES,
the depth of each state variable is unique and can be computed
in polynomial time. Also, it follows that at least one state
variable has depth 0, i.e., no outgoing edges.

\begin{definition}
The depth $d$ of a planning problem $P$ in \THREES{} equals the
largest depth of any state variable $v$ of $P$, i.e.,
$d = \max_{v \in V} d(v)$.
\end{definition}

\noindent
We characterize a planning problem based on the depth of each
of its state variables. Let $n = |V|$ be the number of state
variables, and let $c_i$ denote the number of state variables
with depth $i$. If the planning problem has depth $d$, it
follows that $c_0 + \ldots + c_d = n$. As an example,
consider the planning problem whose causal graph appears in
Figure \ref{fig:3s}. For this planning problem, $n = 8$,
$d = 5$, $c_0 = 2$, $c_1 = 2$, $c_2 = 1$, $c_3 = 1$, $c_4 = 1$,
and $c_5 = 1$.



\begin{lemma}\label{lemma:upperbound}
Consider the values $L_i$ for $i\in \{0, \ldots, d\}$ defined by
$L_d=1$, and $L_i= 2(c_{i+1} L_{i+1} + c_{i+2} L_{i+2} + \ldots + c_d
L_d) + 1$ when $i<d$. The values $L_i$ are an upper bound on the length
of macros generated by our algorithm for a state variable $v$ with
depth $i$.
\end{lemma}

\proof
We prove it by a decreasing induction on the value of $i$.  Assume $v$
has depth $i=d$. It follows from Definition~\ref{def:depth} that $v$
has no incoming edges. Thus, an operator changing the value of $v$ has
no pre-condition on any state variable other than $v$, so $L_d=1$ is
an upper bound, as stated.

Now, assume $v$ has depth $i<d$, and that all $L_{i+k}$ for $k>0$ are
upper bounds on the length of the corresponding macros.  Let $a\in A$
be an operator that changes the value of $v$. From the definition of
depth it follows that $a$ cannot have a pre-condition on a state
variable $u$ with depth $j \leq i$; otherwise there would be an edge
from $u$ to $v$ in the causal graph, causing the depth of $u$ to be
greater than $i$.  Thus, in the worst case, a macro for $v$ has to
change the values of all state variables with depths larger than $i$,
change the value of $v$, and reset the values of state variables at
lower levels. It follows that $L_i = 2(c_{i+1} L_{i+1} + \ldots + c_d
L_d) + 1$ is an upper bound.
\endproof

\begin{theorem}\label{theorem:upperbound}
The upper bounds $L_i$ of Lemma~\ref{lemma:upperbound} satisfy 
$L_i=\Pi_{j=i+1}^d (1+2c_j)$.
\end{theorem}

\proof
Note that
\begin{eqnarray*}
L_i & = & 2(c_{i+1} L_{i+1} + c_{i+2} L_{i+2} + \ldots + c_d L_d) + 1 =\\
& = & 2 c_{i+1} L_{i+1} + 2(c_{i+2} L_{i+2} + \ldots + c_d L_d) + 1 =\\
& = & 2 c_{i+1} L_{i+1} + L_{i+1} = (2 c_{i+1} + 1) L_{i+1}.
\end{eqnarray*}
The result easily follows by induction.
\endproof

Now we can obtain an upper bound $L$ on the total length of the
plan. In the worst case, the goal state assigns a different value to
each state variable than the initial state, i.e.,
$goal(v) \neq init(v)$ for each $v \in V$. To achieve the goal
state the algorithm applies one macro per state variable. Hence
\[
L = c_0 L_0 + c_1 L_1 + \ldots + c_d L_d
= c_0L_0+\frac{L_0-1}{2} = \frac{(1 + 2c_0)L_0 - 1}{2}
= \frac{1}{2}\prod_{j=0}^d(1 + 2c_j) - \frac{1}{2}.
\]

The previous bound depends on the distribution of the variables on depths
according to the causal graph. To obtain a general bound that does not
depend on the depths of the variables we first find which distribution
maximizes the upper bound $L$.

\begin{lemma}\label{lemma:maxL}
The upper bound $L=\frac{1}{2}\prod_{j=0}^d(1 + 2c_j) - \frac{1}{2}$
on planning problems on $n$ variables and depth $d$ is maximized
when all $c_i$ are equal, that is, $c_i=n/(d+1)$.
\end{lemma}

\proof Note that $c_i > 0$ for all $i$, and that
$c_0+\cdots+c_d=n$. The result follows from a direct application of
the well known AM-GM (arithmetic mean-geometric mean) inequality,
which states that the arithmetic mean of positive values $x_i$ is
greater or equal than its geometric mean, with equality only when all
$x_i$ are the same. This implies that the product of positive factors
$x_i=(1+2c_i)$ with fixed sum $A=\sum_{j=0}^d x_j=2n+d$ is maximized
when all are equal, that is, $c_i = n/(d+1)$.
%
%
%
%
\endproof

\begin{theorem}
The length of a plan generated by the algorithm for a planning
problem in \THREES{} with $n$ state variables and depth $d$ is at
most $((1 + 2n/(d+1))^{d+1} - 1)/2$.
\label{thm:len}
\end{theorem}

\proof
This is a direct consequence of Lemma~\ref{lemma:maxL}.  Since $c_0, \ldots,
c_d$ are discrete, it may not be possible to set $c_0 = \ldots = c_d =
n/(d+1)$. Nevertheless, $((1 + 2n/(d+1))^{d+1} - 1)/2$ is an upper
bound on $L$ in this case.\\
\endproof

\noindent
Observe that the bound established in Theorem \ref{thm:len} is an
increasing function of $d$. This implies that for a given $d$, the
bound also applies to planning problems in \THREES{} with depth smaller than
$d$. As a consequence, if the depth of a planning problem in \THREES{} is
bounded from above by $d$, our algorithm generates a solution plan for
the planning problem with polynomial length $O(n^{d+1})$. Since the
complexity of executing a plan is proportional to the plan length, we
can use the depth $d$ to define tractable complexity classes of
planning problems in \THREES{} with respect to plan execution.

\begin{theorem}
The length of a plan generated by the algorithm for a planning
problem in \THREES{} with $n$ state variables is at most
$(3^n - 1)/2$.
\label{thm:maxlen}
\end{theorem}

\proof
In the worst case, the depth $d$ of a planning problem is
$n-1$. It follows from Theorem \ref{thm:len} that the length of a
plan is at most $((1 + 2n/n)^n - 1)/2 = (3^n - 1)/2$.\\
\endproof

\noindent
Note that the bound established in Theorem \ref{thm:maxlen} is
tight; in the second example in Section \ref{sec:examples}, we
showed that our algorithm generates a plan whose length is
$(3^n - 1)/2$.

\begin{lemma}
The complexity of computing the total length of any plan generated by
our algorithm is $O(|V|^2)$.
\label{lemma:plan}
\end{lemma}

\proof
The algorithm generates at most $2|V| = O(|V|)$ macros, 2 for each
state variable. The operator sequence of each macro consists of one
operator and at most $2(|V| - 1) = O(|V|)$ other macros. We can use
dynamic programming to avoid computing the length of a macro more than
once. In the worst case, we have to compute the length of $O(|V|)$
macros, each of which is a sum of $O(|V|)$ terms, resulting in a total
complexity of $O(|V|^2)$.
\endproof

\begin{figure}
\begin{tabular}{ll}
1 & {\bf function} {\sc Operator}($S$, $i$)\\
2 & $o \leftarrow $ first operator in $S$\\
3 & {\bf while} $length(o) < i$ {\bf do}\\
4 & \hspace{15pt} $i \leftarrow i - length(o)$\\
5 & \hspace{15pt} $o \leftarrow $ next operator in $S$\\
6 & {\bf if} $primitive(o)$ {\bf then}\\
7 & \hspace{15pt} {\bf return} $o$\\
8 & {\bf else}\\
9 & \hspace{15pt} {\bf return} {\sc Operator}($o$, $i$)\\
\end{tabular}
\caption{An algorithm for determining the $i$-th operator in a sequence}
\label{fig:opalgo}
\end{figure}

\begin{lemma}
Given a solution plan of length $l$ and an integer $1 \leq i \leq l$,
the complexity of determining the $i$-th operator of the plan is
$O(|V|^2)$.
\label{lemma:op}
\end{lemma}

\proof
We prove the lemma by providing an algorithm for determining the
$i$-th operator, which appears in Figure \ref{fig:opalgo}. Since
operator sequences $S$ consist of operators and macros, the variable
$o$ represents either an operator in $A$ or a macro generated by {\sc
Macro-\THREES}. The function $primitive(o)$ returns $true$ if $o$ is
an operator and $false$ if $o$ is a macro. The function $length(o)$
returns the length of $o$ if $o$ is a macro, and 1 otherwise. We
assume that the length of macros have been pre-computed, which we
know from Lemma \ref{lemma:plan} takes time $O(|V|^2)$.

The algorithm simply finds the operator or macro at the $i$-th
position of the sequence, taking into account the length of macros in
the sequence. If the $i$-th position is part of a macro, the algorithm
recursively finds the operator at the appropriate position in the
operator sequence represented by the macro. In the worst case, the
algorithm has to go through $O(|V|)$ operators in the sequence $S$ and
call {\sc Operator} recursively $O(|V|)$ times, resulting in a total
complexity of $O(|V|^2)$.
\endproof

\subsection{Discussion}

The general view of plan generation is that an output should
consist in a valid sequence of grounded operators that solves a
planning problem. In contrast, our algorithm generates a solution
plan in the form of a system of macros. One might argue that to
truly solve the plan generation problem, our algorithm should
expand the system of macros to arrive at the sequence of
underlying operators. In this case, the algorithm would no longer
be polynomial, since the solution plan of a planning problem in
\THREES{} may have exponential length. In fact, if the only objective
is to execute the solution plan once, our algorithm offers only
marginal benefit over the incremental algorithm proposed by
\citeA{Jonsson98b}.

On the other hand, there are several reasons to view the system
of macros generated by our algorithm as a complete solution to a
planning problem in \THREES. The macros collectively specify all the
steps necessary to reach the goal. The solution plan can be
generated and verified in polynomial time, and the plan can be
stored and reused using polynomial memory. It is even possible to
compute the length of the resulting plan and determine the $i$-th
operator of the plan in polynomial time as shown in Lemmas
\ref{lemma:plan} and \ref{lemma:op}. Thus, for all practical purposes
the system of macros represents a complete solution. Even if the only
objective is to execute the solution plan once, our algorithm should
be faster than that of \citeA{Jonsson98b}. All that is necessary to
execute a plan generated by our algorithm is to maintain a stack of
currently executing macros and select the next operator to execute,
whereas the algorithm of \citeauthor{Jonsson98b} has to perform several
steps for each operator output.

\citeA{Jonsson98b} proved that the bounded plan existence problem
for \THREES{} is $\mathrm{NP}$-hard. The bounded plan existence
problem is the problem of determining whether or not there exists a
valid solution plan of length at most $k$. As a consequence, the
optimal plan generation problem for \THREES{} is $\mathrm{NP}$-hard as
well; otherwise, it would be possible to solve the bounded plan
existence problem by generating an optimal plan and comparing the
length of the resulting plan to $k$. In our examples we have seen
that our algorithm does not generate an optimal plan in general. In
fact, our algorithm is just as bad as the incremental algorithm of
\citeauthor{Jonsson98b}, in the sense that both algorithms may generate
exponential length plans even though there exists a solution of
polynomial length.

Since our algorithm makes it possible to compute the total length of
a valid solution in polynomial time, it can be used to generate
heuristics for other planners. Specifically, \citeA{Katz07} proposed
projecting planning problems onto provably tractable fragments and
use the solution to these fragments as heuristics for the original
problem. We have shown that \THREES{} is such a tractable fragment.
Unfortunately, because optimal planning for \THREES{} is $\mathrm{NP}$-hard,
there is no hope of generating an admissible heuristic. However,
the heuristic may still be informative in guiding the search towards a
solution of the original problem. In addition, for planning problems with
exponential length optimal solutions, a standard planner has no hope
of generating a heuristic in polynomial time, making our macro-based
approach \cite<and that of>{Jonsson07}
the only (current) viable option.

\section{The Class ${\mathbb C}_n$}

\citeA{Domshlak01} defined the class ${\mathbb C}_n$ of planning
problems with multi-valued state variables and chain causal graphs.
Since chain causal graphs are acyclic, it follows that operators
are unary.
Moreover, let $v_i$ be the $i$-th state variable in the chain.
If $i>1$, for each operator $a$ such that $V_{post(a)}\subseteq\{v_i\}$
it holds that $V_{pre(a)} = \{v_{i-1}, v_i\}$.
In other words, each operator that changes the value of a state
variable $v_i$ may only have pre-conditions on $v_{i-1}$ and $v_i$.

The authors showed that there are instances of ${\mathbb C}_n$ with
exponentially sized minimal solutions, and therefore argued that the
class is intractable. In light of the previous section, this argument
on the length of the solutions does not discard the possibility that
instances of the class can be solved in polynomial time using macros.
We show that this is not the case, unless $\mathrm{P}=\mathrm{NP}$.

We define the decision problem \PECn{} as follows. A valid input of
\PECn{} is a planning instance $P$ of ${\mathbb C}_n$. The input $P$
belongs to \PECn{} if and only if $P$ is solvable.  We show in this
section that the problem \PECn{} is $\mathrm{NP}$-hard. This implies
that, unless $\mathrm{P}=\mathrm{NP}$, solving instances of ${\mathbb
C}_n$ is a truly intractable problem, namely, no polynomial-time
algorithm can distinguish between solvable and unsolvable instances of
${\mathbb C}_n$.  In particular, no polynomial-time algorithm can
solve ${\mathbb C}_n$ instances by using macros or any other kind of
output format.\footnote{A valid output format is one that enables
efficient distinction between an output representing a valid plan and
an output representing the fact that no solution was found.}

\begin{figure}
\centerline{\epsfig{figure=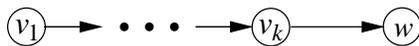, height=0.55cm}}
\vspace{16pt}
\caption{Causal graph of $P(F)$.}
\label{fig:causalCn}
\end{figure}

We prove that \PECn{} is $\mathrm{NP}$-hard by a reduction from
\CNFSAT, that is, the problem of determining whether a CNF formula $F$
is satisfiable.  Let $C_1, \ldots, C_n$ be the clauses of the CNF
formula $F$, and let $v_1, \ldots, v_k$ be the variables that
appear in $F$. We briefly describe the intuition behind the
reduction. The planning problem we create from the formula $F$ has a state
variable for each variable appearing in $F$, and plans are forced
to commit a value (either $0$ or $1$) to these state variables before
actually using them. Then, to satisfy the goal of the problem, these
variables are used to pass messages. However, the operators
for doing this are defined in such a way that a plan can only succeed when
the state variable values it has committed to are a satisfying
assignment of $F$.

We proceed to describe the reduction. First ,we define a planning problem
$P(F)=\langle V, init, goal, A\rangle$ as follows. The set of state variables
is $V = \{v_1, \ldots, v_k, w\}$, where $D(v_i) = \{S, 0, 1, C_1,
C_1', \ldots, C_n, C_n'\}$ for each $v_i$ and $D(w) = \{S, 1, \ldots,
n\}$.  The initial state defines $init(v) = S$ for each $v \in V$ and
the goal state defines $goal(w) = n$. Figure \ref{fig:causalCn} shows
the causal graph of $P(F)$.

\begin{figure}
\centerline{\epsfig{figure=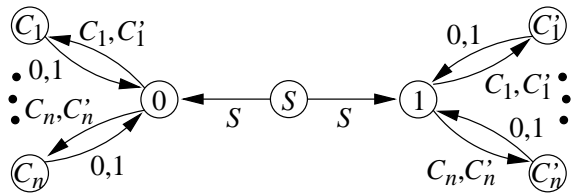, height=2.5cm}}
\vspace{16pt}
\caption{Domain transition graph for $v_i$.}
\label{fig:domaingraph}
\end{figure}

The domain transition graph for each state variable $v_i$ is
shown in Figure \ref{fig:domaingraph}.
Each node represents a value in $D(v_i)$, and an edge from
$x$ to $y$ means that there exists an operator $a$ such that
$pre(a)(v_i) = x$ and $post(a)(v_i)= y$.
Edge labels represent the pre-condition of such operators on
state variable $v_{i-1}$, and multiple labels indicate
that several operators are associated with an edge. We
enumerate the operators acting on $v_i$ using the
notation $a = \langle pre(a); post(a) \rangle$ (when
$i=1$ any mention of $v_{i-1}$ is understood to be void):

\begin{enumerate}
\item[(1)] Two operators $\langle v_{i-1}=S, v_i=S; v_i=0 \rangle$ and
$\langle v_{i-1}=S, v_i=S; v_i=1 \rangle$ that allow $v_i$ to move
from $S$ to either $0$ or $1$.

\item[(2)] Only when $i > 1$. For each clause $C_j$ and each $X \in
\{C_{j}, C_{j}'\}$, two operators \linebreak $\langle v_{i-1}=X,
v_i=0; v_i=C_j \rangle$ and $\langle v_{i-1}=X, v_i=1; v_i=C_j'
\rangle$. These operators allow $v_i$ to move to $C_j$ or $C_j'$ if
$v_{i-1}$ has done so.

\item[(3)] For each clause $C_j$ and each $X\in \{0, 1\}$, an
operator $\langle v_{i-1}=X, v_i=0; v_i=C_j \rangle$ if
$\overline{v}_i$ occurs in clause $C_j$, and an operator $\langle
v_{i-1}=X, v_i=1; v_i=C_j' \rangle$ if $v_i$ occurs in clause $C_j$.
These operators allow $v_i$ to move to $C_j$ or $C_j'$ even if
$v_{i-1}$ has not done so.

\item[(4)] For each clause $C_j$ and each $X=\{0, 1\}$, two operators
$\langle v_{i-1}=X, v_i=C_j; v_i=0 \rangle$ and $\langle v_{i-1}=X,
v_i=C_j'; v_i=1 \rangle$. These operators allow $v_i$ to move back to
$0$ or $1$.
\end{enumerate}
 
\begin{figure}
\centerline{\epsfig{figure=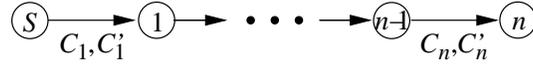,height=0.8cm}}
\vspace{16pt}
\caption{Domain transition graph for $w$.}
\label{fig:domainlast}
\end{figure}

\noindent
The domain transition graph for state variable $w$ is
shown in Figure \ref{fig:domainlast}. For every clause $C_j$ the
only two operators acting on $w$ are $\langle v_k=X, w=j-1; w=j \rangle$,
where $X\in \{C_j, C_j'\}$ (if $j=1$, the pre-condition $w=j-1$ is
replaced by $w=S$).

\begin{proposition} \label{prop:reduction}
A CNF formula $F$ is satisfiable if and only if the planning instance
$P(F)$ is solvable.
\end{proposition}

\proof The proof follows from a relatively straightforward
interpretation of the variables and values of the planning instance
$P(F)$. For every state variable $v_i$, we must use an operator of (1)
to commit to either $0$ or $1$. Note that, once this choice is made,
variable $v_i$ cannot be set to the other value. The reason we need
two values $C_j$ and $C_j'$ for each clause is to enforce this
commitment ($C_j$ corresponds to $v_i = 0$, while $C_j'$ corresponds
to $v_i = 1$). To reach the goal the state variable $w$ has to advance
step by step along the values $1, \ldots, n$. Clearly, for every
clause $C_j$ there must exist some variable $v_i$ that is first set to
values $C_j$ or $C_j'$ using an operator of (3). Then, this
``message'' can be propagated along variables $v_{i+1}, \ldots, v_{k}$
using operators of (2).  Note that the existence of an operator of (3)
acting on $v_i$ implies that the initial choice of $0$ or $1$ for
state variable $v_i$, when applied to the formula variable $v_i$,
makes the clause $C_j$ true. Hence, if $\Pi$ is a plan solving $P(F)$,
we can use the initial choices of $\Pi$ on state variables $v_i$ to
define a (partial) assignment $\sigma$ that satisfies all clauses of
$F$.

Conversely, if $\sigma$ is some assignment that satisfies $F$, we show
how to obtain a plan $\Pi$ that solves $P(F)$. First, we set every
state variable $v_i$ to value $\sigma(v_i)$. For every one of the
clauses $C_j$, we choose a variable $v_i$ among those that make $C_j$
true using assignment $\sigma$. Then, in increasing order of $j$, we
set the state variable $v_i$ corresponding to clause $C_j$ to a value
$C_j$ or $C_j'$ (depending on $\sigma(v_i)$), and we pass this message
along $v_{i+1},\ldots,v_k$ up to $w$.
\endproof

\begin{theorem} \PECn{} is $\mathrm{NP}$-hard.
\end{theorem}

\proof
Producing a planning instance $P(F)$ from a CNF formula $F$
can be easily done in polynomial time, so we have a polynomial-time
reduction \CNFSAT{} $\le_p$ \PECn.
\endproof

\section{Polytree Causal Graphs}

In this section, we study the class of planning problems with binary
state variables and polytree causal graphs.  \citeA{Brafman03} presented
an algorithm that finds plans for problems of this class in time
$O(n^{2\kappa})$, where $n$ is the number of variables and $\kappa$ is
the maximum indegree of the polytree causal graph. \citeA{Brafman06}
also showed how to solve in time roughly $O(n^{\omega\delta})$ planning
domains with local depth $\delta$ and causal graphs of tree-width
$\omega$. It is interesting to observe that both algorithms fail to
solve polytree planning domains in polynomial time for different
reasons: the first one fails when the tree is too \emph{broad}
(unbounded indegree), the second one fails when the tree is too
\emph{deep} (unbounded local depth, since the tree-width $\omega$ of a
polytree is 1).

\begin{figure}
\centerline{\epsfig{figure=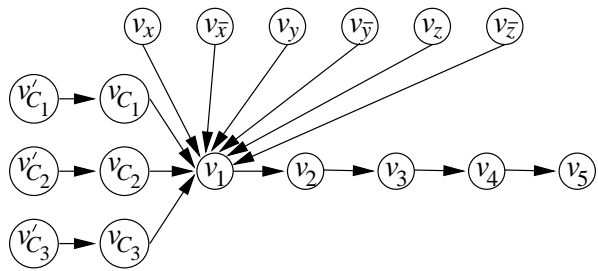,height=3.5cm}}
\vspace{16pt}
\caption{Causal graph of $P_F$ when $F=C_1\wedge C_2\wedge C_3$ on three
variables $x, y, z$.}
\label{fig:reduction}
\end{figure}

In this section we prove that the problem of plan existence for
polytree causal graphs with binary variables is
$\mathrm{NP}$-hard. Our proof is a reduction from \THREESAT{} to this
class of planning problems. As an example of the reduction,
Figure~\ref{fig:reduction} shows the causal graph of the planning
problem $P_F$ that corresponds to a formula $F$ with three variables
and three clauses (the precise definition of $P_F$ is given in
Proposition~\ref{prop:reduction2}). Finally, at the end of this
section we remark that the same reduction solves a problem expressed
in terms of CP-nets \cite{Boutilier04}, namely, that dominance testing
for polytree CP-nets with binary variables and partially specified
CPTs is $\mathrm{NP}$-complete.

Let us describe briefly the idea behind the reduction. The planning
problem $P_F$ has two different parts. The first part (state variables
$v_{x}, v_{\overline{x}}, \ldots, v_{C_1}, v'_{C_1}, \ldots,$ and
$v_1$) depends on the formula $F$ and has the property that a plan
may change the value of $v_1$ from $0$ to $1$ as many times as the
number of clauses of $F$ that a truth assignment can satisfy. However,
this condition on $v_1$ cannot be stated as a planning problem
goal. We overcome this difficulty by introducing a second part (state
variables $v_1, v_2,\ldots,v_t$) that translates it to a regular planning
problem goal.

We first describe the second part. Let $P$ be the planning problem
$\langle V, init, goal, A \rangle$ where $V$ is the set of state
variables $\{v_1, \ldots, v_{2k-1}\}$ and $A$ is the set of
$4k-2$ operators $\{\alpha_1,\ldots, \alpha_{2k-1},\beta_1,\ldots,
\beta_{2k-1}\}$. For $i = 1$, the operators are defined as
$\alpha_1 = \langle v_1 = 1; v_1 = 0 \rangle$ and
$\beta_1 = \langle v_1 = 0; v_1 = 1 \rangle$.
For $i > 1$, the operators are
$\alpha_i = \langle v_{i-1}=0, v_i = 1; v_i = 0 \rangle$ and
$\beta_i = \langle v_{i-1}=1, v_i = 0; v_i = 1\rangle$. The initial
state is $init(v_i) = 0$ for all $i$, and the goal state is
$goal(v_i) = 0$ if $i$ is even and $goal(v_i) = 1$ if odd.

\begin{lemma}\label{lemma:P}
Any valid plan for planning problem $P$ changes state variable $v_1$
from 0 to 1 at least $k$ times. There is a valid plan that
achieves this minimum.
\end{lemma}

\proof Let $A_i$ and $B_i$ be, respectively, the sequences of
operators $\langle \alpha_1,\ldots, \alpha_i \rangle$ and $\langle
\beta_1, \ldots, \beta_i \rangle$. It is easy to verify that the plan
$\langle B_{2k-1}, A_{2k-2}, B_{2k-3} , \ldots, B_3, A_2, B_1 \rangle$
solves the planning problem $P$. Indeed, after applying the operators
of $A_{i}$ (respectively, the operators of $B_{i}$), variables $v_1,
\ldots, v_i$ become $0$ (respectively, $1$). In particular, variable
$v_i$ attains its goal state ($0$ if $i$ is even, $1$ if $i$ is
odd). Subsequent operators in the plan do not modify $v_i$, so the
variable remains in its goal state until the end. The operator
$\beta_1$ appears $k$ times in the plan (one for each sequence of type
$B_i$), thus the value of $v_1$ changes $k$ times from $0$ to $1$.

We proceed to show that $k$ is the minimum. Consider some plan $\Pi$
that solves the planning problem $P$, and let $\lambda_i$ be the
number of operators $\alpha_i$ and $\beta_i$ appearing in $\Pi$ (in
other words, $\lambda_i$ is the number of times that the value of
$v_i$ changes, either from $0$ to $1$ or from $1$ to $0$). Note that
the number of times operator $\beta_i$ appears is equal to or
precisely one more than the number of occurrences of $\alpha_i$. We
will show that $\lambda_{i-1}>\lambda_{i}$. Since $\lambda_{2k-1}\leq
1$, this implies that $\lambda_1\geq 2k-1$, so that plan $\Pi$ has, at
least, $k$ occurrences of $\beta_1$, completing the proof.

We show that $\lambda_{i-1}>\lambda_{i}$. Let $S_i$ be the subsequence
of operators $\alpha_i$ and $\beta_i$ in plan $\Pi$. Clearly, $S_i$
starts with $\beta_i$ (since the initial state is $v_i=0$), and the
same operator cannot appear twice consecutively in $S_i$, so
$S_i=\beta_i, \alpha_i, \beta_i, \alpha_i$, etc.  Also note that, for
$i>1$, $\beta_i$ has $v_{i-1}=1$ as a pre-condition, and $\alpha_i$
has $v_{i-1}=0$, hence there must be at least one operator
$\alpha_{i-1}$ in plan $\Pi$ betweeen any two operators $\beta_i$ and
$\alpha_i$. For the same reason we must have at least one operator
$\beta_{i-1}$ between any two operators $\alpha_i$ and $\beta_i$, and
one operator $\beta_{i-1}$ before the first operator $\beta_i$. This
shows that $\lambda_{i-1}\geq\lambda_{i}$. On the other hand, variables $v_i$
and $v_{i-1}$ have different values in the goal state, so subsequences $S_i$ and
$S_{i-1}$ must have different lengths, that is, $\lambda_{i-1}\neq
\lambda_i$. Together, this implies $\lambda_{i-1}>\lambda_i$, as desired.
\endproof

\begin{proposition}\label{prop:reduction2}
\THREESAT{} reduces to plan existence for planning problems with binary
variables and polytree causal graphs.
\end{proposition}

\proof
Let $F$ be a CNF formula with $k$ clauses and $n$ variables. We produce
a planning problem $P_F$ with $2n+4k-1$ state variables and
$2n+14k-3$ operators.  The planning problem has two state variables $v_x$
and $v_{\overline{x}}$ for every variable $x$ in $F$, two state variables
$v_C$ and $v_C'$ for every clause $C$ in $F$, and $2k-1$ additional
variables $v_1,\ldots,v_{2k-1}$. All variables are $0$ in the initial
state. The (partial) goal state is defined by
$V_{goal} = \{v_1, \ldots, v_{2k-1}\}$, $goal(v_i)=0$ when $i$ is even,
and $goal(v_i)=1$ when $i$ is odd, like in problem $P$ of
Lemma \ref{lemma:P}. The operators are:

\begin{enumerate}
\item[(1)] Operators $\langle v_x = 0; v_x = 1 \rangle$ and
$\langle v_{\overline{x}} = 0; v_{\overline{x}} = 1 \rangle$ for
every variable $x$ of $F$.
\item[(2)] Operators $\langle v_C' = 0; v_C' = 1 \rangle$,
$\langle v_C'= 0, v_C = 0; v_C = 1 \rangle$ and
$\langle v_C'= 1, v_C = 1; v_C = 0 \rangle$ for every clause $C$ of $F$.
\item[(3)] Seven operators for every clause $C$, one for each
partial assignment that satisfies $C$.
Without loss of generality, let $x$, $y$, and $z$ be the three variables
that appear in clause $C$.
Then for each operator $a$ among these seven, $V_{pre(a)} = \{v_x,
v_{\overline{x}}, v_y, v_{\overline{y}}, v_z, v_{\overline{z}}, v_C, v_1\}$,
$V_{post(a)} = \{v_1\}$, $pre(a)(v_C) = 1$, $pre(a)(v_1) = 0$, and
$post(a)(v_1) = 1$.
The pre-condition on state variables
$v_x, v_{\overline{x}}, v_y, v_{\overline{y}}, v_z, v_{\overline{z}}$
depends on the corresponding satisfying partial assignment.
For example, the operator corresponding to the partial
assignment $\{x = 0, y = 0, z = 1\}$ of the clause
$C=x\vee \overline{y} \vee \overline{z}$
has the pre-condition $(v_x = 0, v_{\overline{x}} = 1, v_y = 0,
v_{\overline{y}} = 1, v_z = 1, v_{\overline{z}} = 0).$
\item[(4)] An operator
$\langle (\forall C, v_C = 0), v_1 = 1; v_1 = 0 \rangle$.
\item[(5)] Operators $\alpha_i=\langle v_{i-1}=0, v_i = 1; v_i = 0 \rangle$
and $\beta_i=\langle v_{i-1}=1, v_i = 0; v_i = 1 \rangle$ for
$2\leq i \leq 2k-1$ (the same operators as in problem $P$ except for
$\alpha_1$ and $\beta_1$).
\end{enumerate}

\noindent
We note some simple facts about problem $P_F$. For any variable $x$, state
variables $v_x$ and $v_{\overline{x}}$ in $P_F$ start at $0$, and by applying
the operators in (1) they can change into $1$ but not back to $0$.
In particular, a plan $\Pi$ cannot reach both of the partial states
$\langle v_x=1, v_{\overline{x}}=0 \rangle$ and
$\langle v_x=0, v_{\overline{x}}=1 \rangle$ during
the course of its execution.

Similarly, if $C$ is a clause of $F$, state variable $v_C$ can change from $0$
to $1$ and, by first changing $v_C'$ into $1$, $v_C$ can change back to $0$.
No further changes are possible, since no operator brings back $v_C'$ to $0$.

Now we interpret operators in (3) and (4), which are the only operators
that affect $v_1$. To change $v_1$ from $0$ to $1$ we need to apply
one of the operators in (3), thus we require $v_C=1$ for a clause
$C$. But the only way to bring back $v_1$ to $0$ is applying the
operator in (4) which has as pre-condition that $v_C=0$. We deduce that
every time that $v_1$ changes its value from $0$ to $1$ and then back
to $0$ in plan $\Pi$, at least one of the $k$ state variables $v_C$
is \emph{used up}, in the sense that $v_C$ has been brought from $0$ to
$1$ and then back to $0$, and cannot be used again for the same purpose.

We show that $F$ is in \THREESAT{} if and only if there is a valid
plan for problem $P_F$. Assume $F$ is in \THREESAT{}, and let $\sigma$
be a truth assignment that satisfies $F$. Consider the following plan
$\Pi'$. First, we set $v_x=\sigma(x)$ and
$v_{\overline{x}}=1-\sigma(x)$ for all variables $x$ using the
operators of (1). Then, for a clause $C$ in $F$, we set $v_C=1$, we
apply the operator of (3) that corresponds to $\sigma$ restricted to
the variables of clause $C$ (at this point, $v_1$ changes from $0$ to
$1$), then we set $v_C'=1$ and $v_C=0$, and we apply the operator of
(4) (at this point, $v_1$ change from $1$ to $0$). By repeating this
process for every clause $C$ of $F$ we are switching the state
variable $v_1$ exactly $k$ times from $0$ to $1$. Now, following the
proof of Lemma~\ref{lemma:P}, we can easily extend this plan $\Pi'$ to
a plan $\Pi$ that sets all variables $v_i$ to their goal values.

We show the converse, namely, that the existence of a valid plan $\Pi$
in $P_F$ implies that $F$ is satisfiable. Define an assignment
$\sigma$ by setting $\sigma(x)=1$ if the partial state $\{v_x=1,
v_{\overline{x}}=0\}$ appears during the execution of $\Pi$, and
$\sigma(x)=0$ otherwise. (Recall that at most one of the partial
states $\{v_x=1, v_{\overline{x}}=0\}$ and $\{v_x=0,
v_{\overline{x}}=1\}$ can appear during the execution of any
plan). By Lemma~\ref{lemma:P}, $\Pi$ must be such that state variable
$v_1$ changes from $0$ to $1$ at least $k$ times. This implies that
$k$ operators of (3), all of them corresponding to different clauses,
have been used to move $v_1$ from $0$ to $1$. But to apply such an
operator, the values of state variables $\{v_x, v_{\overline{x}}\}$
must satisfy the corresponding clause. Thus the assignment $\sigma$
satisfies all the $k$ clauses of $F$.

\endproof

\begin{theorem}\label{thm:reduction}
Plan existence for planning problems with binary variables and
polytree causal graph is $\mathrm{NP}$-complete.
\end{theorem}

\proof Due to Proposition~\ref{prop:reduction2} we only need to show
that the problem is in $\mathrm{NP}$. But \citeA{Brafman03} showed
that this holds in the more general setting of planning problems with
causal graphs where each component is directed-path singly connected
(that is, there is at most one directed path between any pair of
nodes). Their proof exploits a non-trivial auxiliary result: solvable
planning problems on binary variables with a directed-path singly
connected causal graph have plans of polynomial length (the same is
not true for non-binary variables, or unrestricted causal graphs).
\endproof

\subsection{CP-nets}

\citeA{Boutilier04} introduced the notion of a CP-net as a graphical
representation of user preferences. In brief, a CP-net is a network of
dependences on a set of variables: the preferences the user has for a
variable depend on the values of some of the others, under the
\emph{ceteris paribus} (all else being equal) assumption, that is, the
user preferences on the variable are completely independent of the
values of the variables not mentioned. The preferences for a variable
given its parent variables in the network are stored in conditional
preference tables, or CPTs.

\citeA{Boutilier04} showed that the \emph{dominance query} problem in
acyclic CP-nets, that is, the problem of deciding if one variable
outcome is preferable to another, can be expressed in terms of a
planning problem. The network of dependences of the CP-net becomes the
causal graph of the planning problem.

However, under certain conditions, we can perform the opposite process:
transform a planning problem into a CP-net and a dominance query
problem, such that answering the query amounts to solving the planning
problem. This is possible under the following conditions on planning
problems with acyclic causal graph and binary variables:

\begin{enumerate}
\item Two operators that modify the same variable in opposing
directions must have non-matching prevail conditions (the prevail
condition of an operator $a$ is the partial state $pre(a) \mid
V - V_{post(a)}$).
\item We must allow partially specified CPTs in the CP-net description.
\end{enumerate}

\noindent
The first condition guarantees that we obtain consistent CPTs
from the planning instance operators. The second condition ensures that
the reduction is polynomial-size preserving, since fully specified CPTs
are exponential in the maximum node indegree of the CP-net.

In particular, the planning instance $P_F$ we reduced $F$ to satisfies
the first condition. (Note that this is not true for the planning
problem $P$ of Lemma~\ref{lemma:P}, but we drop the reversing
operators $\alpha_1$ and $\beta_1$ when constructing $P_F$ in
Proposition~\ref{prop:reduction2}.) As a consequence, we can claim the
following:

\begin{theorem}\label{thm:reductionCP}
Dominance testing for polytree CP-nets with binary variables and
partially specified CPTs is $\mathrm{NP}$-complete.
\end{theorem}

\section{Conclusion}

We have presented three new complexity results for planning problems
with simple causal graphs. First, we provided a polynomial-time
algorithm that uses macros to generate solution plans for the class
\THREES. Although the solutions are generally suboptimal, the algorithm
can generate representations of exponentially long plans in polynomial
time. This has several implications for theoretical work in planning,
since it has been generally accepted that exponentially sized minimal
solutions imply that plan generation is intractable. Our work shows
that this is not always the case, provided that one is allowed to
express the solution in a succinct notation such as macros.  We also
showed that plan existence for the class ${\mathbb C}_n$ is
$\mathrm{NP}$-hard, and that plan existence for the class of planning
problems with binary variables and polytree causal graph is
$\mathrm{NP}$-complete.

\citeA{Jonsson98b} investigated whether plan generation is
significantly harder than plan existence. Using the class \THREES, they
demonstrated that plan existence can be solved in polynomial time,
while plan generation is intractable in the sense that solution plans
may have exponential length. Our work casts new light on this result:
even though solution plans have exponential length, it is possible to
generate a representation of the solution in polynomial time.  Thus,
it appears as if for the class \THREES, plan generation is not inherently
harder than plan existence. We are not aware of any other work that
determines the relative complexity of plan existence and plan
generation, so the question of whether plan generation is harder that
plan existence remains open.

A potential criticism of our algorithm is that a solution in the form
of macros is not standard, and that it is intractable to expand the
system of macros to arrive at the possibly exponentially long sequence
of underlying operators. Although this is true, we have shown that the
system of macros share several characteristics with a proper solution.
It is possible to generate and validate the solution in polynomial
time, and the solution can be stored using polynomial memory. We also
showed that it is possible to compute the total length of the solution
in polynomial time, as well as determine which is the $i$-th operator
in the underlying sequence.

Since they are relatively simple, the class ${\mathbb C}_n$ and the
class of planning problems with binary state variables and polytree
causal graphs could be seen as promising candidates for proving the
relative complexity of plan existence and plan generation. However,
we have shown that plan existence for ${\mathbb C}_n$ is
$\mathrm{NP}$-hard, and that plan existence for planning problems
with polytree causal graphs is $\mathrm{NP}$-complete. Consequently,
these classes cannot be used to show that plan generation is
harder than plan existence, since plan existence is already
difficult. Our work also closes the complexity gaps that
appear in the literature regarding these two classes.

It is however possible that there exist subsets of planning problems
in these classes for which plan existence can be solved in polynomial
time. In fact, for polytree causal graphs in binary variables we know
that this is the case, due to the algorithms of \citeA{Brafman03,Brafman06}
mentioned in Section~6. Hence the plan generation
problem is polynomial if we restrict to polytree causal graphs with
either bounded indegree $\kappa$  or bounded local depth $\delta$.
Consequently, our reduction from \THREESAT{} exhibits both unbounded
indegree and unbounded local depth.

Similarly, one may ask if the class ${\mathbb C}_n$ of planning
problems has some parameter that, when bounded, would yield a
tractable subclass.  The state variables in our reduction have domains
whose size depends on the number of clauses of the corresponding CNF
formula, so the domain size appears as an interesting
candidate. Planning problems of ${\mathbb C}_n$ with binary variables
are tractable due to the work of \citeA{Brafman03}, but the ideas they
use do not extend to domain sizes other than 2.  Hence it would be
interesting to investigate whether the problem of plan existence for
the class ${\mathbb C}_n$ is easier if the size of the state variable
domains is bounded by a constant.

\begin{appendix}
\section{Proof of Theorem \ref{thm:3s}}
\label{app:thm}

Assume that {\sc GenerateMacro}($P$, $v$, $x$, $M$) successfully
returns the macro $m_x^v = \langle S_1, a, S_0 \rangle$. Let $U = \{u
\in V_{pre(a)} - \{v\} \mid pre(a)(u) = 1\}$ and let $W = \{w_1,
\ldots, w_k\} \subseteq U$ be the set of state variables in $U$ such
that $w_i$ is not splitting, $\{m_0^{w_i}, m_1^{w_i}\} \in M$, and
$w_i$ comes before $w_j$ in topological order if and only if $i <
j$. It follows that no $u \in U$ is static, that $S_1 = \langle
m_1^{w_k}, \ldots, m_1^{w_1} \rangle$ and that $S_0 = \langle
m_0^{w_1}, \ldots, m_0^{w_k} \rangle$. Since each state variable $w_i
\in W$ is not splitting, it has to be symmetrically reversible.

\begin{lemma}
\label{lem:preOfAncestors}
For each $w_i \in W$, $pre^{w_i} \sqsubseteq pre^v$.
\end{lemma}

\proof
Since $w_i \in V_{pre(a)}$ and $v \in V_{post(a)}$, there is an edge
from $w_i$ to $v$ in the causal graph.  Thus, any ancestor of $w_i$ is
also an ancestor of $v$, so $Anc^{w_i} \subset Anc^v$. For a state
variable $u \in Anc^{w_i}$, $pre^{w_i}(u) = 1$ if and only if $u$ is
splitting and $w_i \in V_1^u$. The graph $G_1^u = (V, E_1^u)$ includes
the edge from $w_i$ to $v$, which means that $v \in V_1^u$ if and only
if $w_i \in V_1^u$. It follows that $pre^{w_i}(u) = 1$ if and only if
$pre^v(u) = 1$, and as a consequence, $pre^{w_i} \sqsubseteq pre^v$.\\
\endproof

Let $\Pi=\langle S_0, a, S_1\rangle$. For each $w_i \in W$ and $y \in
\{0,1\}$, let $\Pi^{w_i}_y$ be the sequence preceding the macro
$m^{w_i}_y$ in $\Pi$, that is, $\Pi^{w_i}_1 = \langle m_1^{w_k},
\ldots, m_1^{w_{i+1}} \rangle$ and $\Pi^{w_i}_0 = \langle S_0, a,
m_0^{w_1}, \ldots, m_0^{w_{i-1}} \rangle$. Further, let $\Pi^a$ be the
sequence appearing before $a$, that is, $\Pi^a =
\langle S_0 \rangle$.

\begin{lemma}
\label{lem:post}
For each $1\leq i \leq k$, the post-conditions of sequences
$\Pi^{w_i}_1$, $\Pi^a$, and $\Pi^{w_i}_0$ are
\begin{itemize}
\item $post(\Pi^{w_i}_1)=(w_{i+1}=1, \ldots, w_k=1)$,
\item $post(\Pi^a)=(w_1=1, \ldots, w_k=1)$,
\item $post(\Pi^{w_i}_0)=(w_1=0, \ldots, w_{i-1}=0, w_{i}=1, \ldots, w_k=1, v=x)$. 
\end{itemize}
\end{lemma}

\proof A direct consequence of $post(\langle a_1, \ldots,
a_k\rangle)=post(a_1)\oplus \cdots \oplus post(a_k)$ and
$post(m^{w_i}_y)=(w_i=y)$, $post(a)=(v=x)$.  \endproof

\begin{lemma}
\label{lem:pre}
For each $1\leq i \leq k$, the pre-conditions of sequences
$\Pi^{w_i}_1$, $\Pi^a$, $\Pi^{w_i}_0$, and $\Pi$ satisfy
$pre(\Pi^{w_i}_1) \sqsubseteq pre(\Pi^a) \sqsubseteq pre(\Pi^{w_i}_0)
\sqsubseteq pre(\Pi) \sqsubseteq pre^v\oplus (v=1-x)$.
\end{lemma}

\proof
Since $pre(\langle a_1, \ldots, a_k \rangle)=pre(a_k)\oplus \cdots
\oplus pre(a_1)$, it follows that $pre(\Pi^{w_i}_1)\sqsubseteq
pre(\Pi^{a}) \sqsubseteq pre(\Pi^{w_i}_0) \sqsubseteq pre(\Pi)$. We
prove that $pre(\Pi)\sqsubseteq pre^v\oplus (v=1-x)$. For
a state variable $u$ such that $pre(\Pi)(u)\neq \perp$, let $m^u$ be
the first operator in $\langle S_0, a, S_1\rangle$ such that $u\in
V_{pre(m^u)}$, so that $pre(\Pi)(u)=pre(m^u)(u)$. 

If $m^u=m^{w_i}_1$, then it follows that $pre(m^u)\sqsubseteq
pre^{w_i}\oplus (w_i=0)\sqsubseteq pre^v$, where we have used that
$m^{w_i}_1$ is a \THREES-macro, $w_i$ is symmetrically reversible, and
that $pre^{w_i}\sqsubseteq pre^v$ due to
Lemma~\ref{lem:preOfAncestors}. In particular, $pre(m^u)(u)=pre^v(u)$.

Since we assume that planning problems are in normal form, $u=w_i$
implies that $u\in V_{pre(m^{w_i}_1)}$. It follows that if $m^u\neq
m^{w_i}_1$ for all $i$, then $u\neq w_i$ for all $i$. If
$m^u=m^{w_i}_0$ we have that $pre(m^u)\sqsubseteq pre^{w_i}\oplus
(w_i=1)$, but due to $u\neq w_i$, we deduce
$pre(m^u)(u)=pre^{w_i}(u)=pre^{v}(u)$.

Finally, consider the case $m^u=a$. If $u=v$ then $pre(m^u)(u)=1-x$,
as desired. If $u\neq v$ is splitting, then either $v$ belongs to
$V_0^u$ and $pre(m^u)(u)=0$, or $v$ belongs to $V_1^u$ and
$pre(m^u)(u)=1$. That is, $pre(m^u)(u)=pre^v(u)$. If $u\neq v$ is
symmetrically reversible it follows that $pre(m^u)(u)=0$, since the
case $pre(m^u)(u)=1$ would have forced the algorithm to either fail or
include $u$ in $W$. If $u \neq v$ is static, $pre(m^u)(u)=0$, else
the algorithm would have failed.
\endproof

\begin{lemma}
\label{lem:helper}
Let $p, p', q$ and $r$ be partial states. If $p \sqsubseteq p'$ and
$(p'\oplus q)\triangledown r$, then $(p \oplus q)\triangledown r$.
\end{lemma}

\proof A direct consequence of $p \oplus q \sqsubseteq p'\oplus q$.
\endproof

\begin{lemma} The macro $m^v_x$ generated by the algorithm is well-defined.
\label{lem:welldefined}
\end{lemma}

\proof
Since $\Pi$ only includes macros for the ancestors of $v$ in the
causal graph, and since the causal graph is acyclic, no cyclic
definitions occur. It remains to show that, for a macro $m$ in $\Pi$
and a sequence $\Pi^m$ preceding $m$ in $\Pi$, it holds that
$(pre(\Pi^m)\oplus post(\Pi^m)) \triangledown pre(m)$. Note that due
to Lemmas~\ref{lem:pre} and \ref{lem:helper} it is enough to show that
\begin{enumerate}
\item[(a)] $(pre^v \oplus (v=1-x)\oplus post(\Pi^{w_i}_1)) \triangledown pre(m^{w_i}_1),$
\item[(b)] $(pre^v \oplus (v=1-x)\oplus post(\Pi^a)) \triangledown
pre(a),$
\item[(c)] $(pre^v \oplus (v=1-x)\oplus post(\Pi^{w_i}_0)) \triangledown pre(m^{w_i}_0).$
\end{enumerate}

Case (a) follows easily since $V_{post(\Pi^{w_i}_1)} \cap
V_{pre(m^{w_i}_1)} = \emptyset$ and $pre(m^{w_i}_1)=pre^{w_i}\oplus
(w_i=0) \sqsubseteq \linebreak pre^v$. Case (c) is similar, although
this time we must use that $post(\Pi^{w_i}_0)(w_i)=1$ and
$post(\Pi^{w_i}_0)(w_j)=0$ for $j<i$, as required by
$pre(m^{w_i}_0)=pre^{w_i}\oplus (w_i=1)$. Finally, case (b) holds
because a variable $u \in V_{pre(a)}$ can be either $u=v$, which is
covered by $(v=1-x)$, splitting or static, which is covered by
$pre^v$, or symmetrically reversible, which is covered by $pre^v(u)=0$
if $pre(a)(u)=0$, and by $post(\Pi^a)(u)=1$ if $pre(a)(u)=1$.\\
\endproof

In remains to show that $m^v_x$ is a \THREES-macro.  It follows from
Lemmas~\ref{lem:pre} and \ref{lem:welldefined} that it is well-defined
and it satisfies $pre(m^v_x)=pre(\Pi)\sqsubseteq pre^v\oplus
(v=1-x)$. Finally, $post(m^v_x)=post(\Pi)-pre(\Pi)=(v=x)$ is a direct
consequence of $post(\Pi)=(w_1=0, \ldots, w_k=0, v=x)$ from
Lemma~\ref{lem:post}, and $pre(\Pi)(w_i)=0$, $pre(\Pi)(v)=1-x$ from
the proof of Lemma~\ref{lem:pre}.

\end{appendix}

\acks{This work was partially funded by MEC grants TIN2006-15387-C03-03 and
TIN2004-07925-C03-01 (GRAMMARS).}

\bibliography{gimenez08a}
\bibliographystyle{theapa}

\end{document}